\documentclass[twoside,journal]{IEEEtran}

\usepackage{adjustbox}
\usepackage{booktabs}
\usepackage{subcaption}
\usepackage{gensymb}
\usepackage{xcolor}
\usepackage{epsfig}
\usepackage{graphicx}
\usepackage{amsmath}
\usepackage{amssymb}
\usepackage{textcomp}
\usepackage{multirow}
\usepackage{float}
\usepackage{pifont}
\definecolor{darkgreen}{RGB}{0, 150, 0}
\definecolor{darkred}{RGB}{200, 0, 0}
\usepackage[colorlinks,pagebackref=false,citecolor=blue,bookmarks=false,hypertexnames=true]{hyperref}

\makeatletter
\def\mythanks{\gdef\@thefnmark{}\@footnotetext}
\makeatother

\newcommand{\rd}[1]{\textcolor{black}{#1}}

\newcommand{\ch}{{\color{darkgreen} \ding{51}}}
\newcommand{\xm}{{\color{darkred} \ding{55}}}

\newcommand\sizeBEV{.156}
\newcommand\sizeFish{.2075}

\begin{document}
\title{SynWoodScape: Synthetic Surround-view Fisheye Camera Dataset for Autonomous Driving}

\author{\large 
Ahmed Rida Sekkat, 
Yohan Dupuis, 
Varun Ravi Kumar, 
Hazem Rashed, \\
\large 
Senthil Yogamani,
Pascal Vasseur, and 
Paul Honeine
}

\twocolumn[{
  \captionsetup{singlelinecheck=false, font=footnotesize, belowskip=-2pt}
	\renewcommand\twocolumn[1][]{#1}
	\maketitle
	\begin{center}
		\vspace{-0.6cm}
		\includegraphics[width=0.85\textwidth]{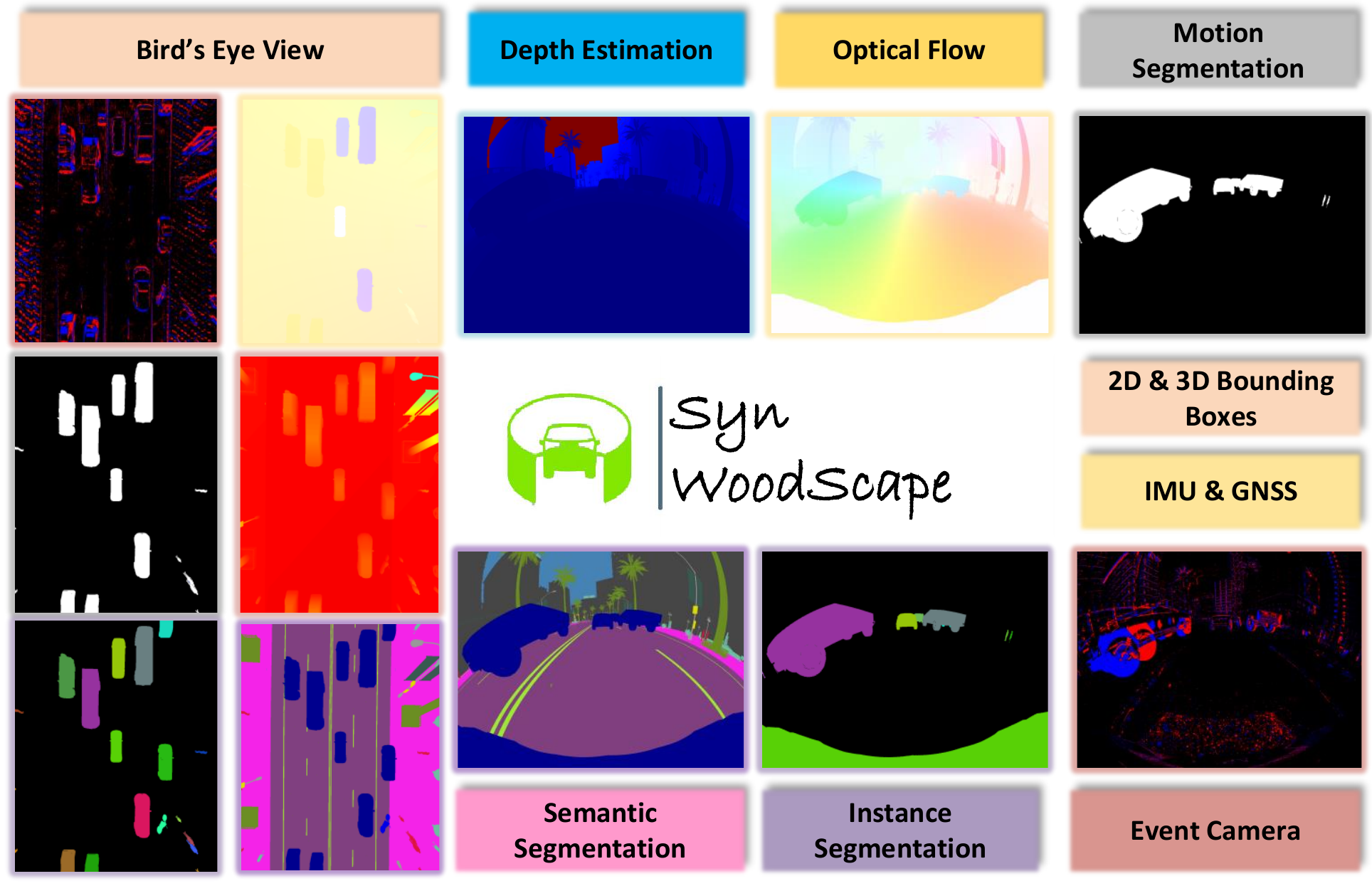}
		\captionof{figure}
		{
            \bf Overview of all the SynWoodScape tasks. The dataset and the baseline code will be released in \url{https://woodscape.valeo.com}.
		}
		\label{fig:teaser}
	\end{center}
}]
\mythanks{A. R. Sekkat and P. Honeine are with Université de Rouen Normandie, LITIS Lab, Rouen, France.}
\mythanks{V. Ravi Kumar, H. Rashed are with Valeo DAR, Kronach, Germany.}
\mythanks{S. Yogamani is with Valeo Vision Systems, Tuam, Ireland.}
\mythanks{Y. Dupuis is with LINEACT CESI, Paris La Défense, France.}
\mythanks{P. Vasseur is with MIS Lab
, Universit\'{e} de Picardie Jules Verne, France.}%
\mythanks{This research has been partially funded by the ANR Project CLARA ANR-18-CE33-0004.}
\mythanks{The authors would like to thank Nawal Bouin and Pierre-Sylvain Luquet from Normandie Valorisation for their support.}

\begin{abstract}

Surround-view cameras are a primary sensor for automated driving, used for near-field perception. It is one of the most commonly used sensors in commercial vehicles primarily used for parking visualization and automated parking. Four fisheye cameras with a $190^\degree$ field of view cover the $360^\degree$ around the vehicle. Due to its high radial distortion, the standard algorithms do not extend easily. Previously, we released the first public fisheye surround-view dataset named WoodScape. In this work, we release a synthetic version of the surround-view dataset, covering many of its weaknesses and extending it. Firstly, it is not possible to obtain ground truth for pixel-wise optical flow and depth. Secondly, WoodScape did not have all four cameras annotated simultaneously in order to sample diverse frames. However, this means that multi-camera algorithms cannot be designed to obtain a unified output in birds-eye space, which is enabled in the new dataset. We implemented surround-view fisheye geometric projections in CARLA Simulator matching WoodScape's configuration and created SynWoodScape. We release 80k images from the synthetic dataset with annotations for 10+ tasks\footnote{An initial sample of the dataset is released in \href{https://drive.google.com/drive/folders/1N5rrySiw1uh9kLeBuOblMbXJ09YsqO7I}{link}. }. We also release the baseline code and supporting scripts.
\end{abstract}
\begin{IEEEkeywords}
Fisheye Cameras, Omnidirectional vision, Automated Driving, Synthetic Datasets.
\end{IEEEkeywords}
\section{Introduction}

In autonomous driving (AD), the near field is a region from 0 to 30 meters and $360^\degree$ coverage around the vehicle. Near-field perception is primarily needed for use cases, such as automated parking, traffic jam assist, and urban driving, where the predominant sensor suite includes surround-view fisheye-cameras and ultrasonics \cite{popperli2019capsule}. Despite the importance of such use cases, most research to date has focused on far-field perception. Consequently, there are limited datasets and research papers on near-field perception tasks. In contrast to far-field, near-field perception is more challenging due to high precision object detection requirements of 10 cm \cite{eising2021near}. For example, an autonomous car needs to be parked in a tight space where high precision detection is required with no room for error.\par

\begin{figure*}
  \captionsetup{singlelinecheck=false, font=footnotesize, belowskip=-14pt}
  \centering
    \includegraphics[width=\linewidth]{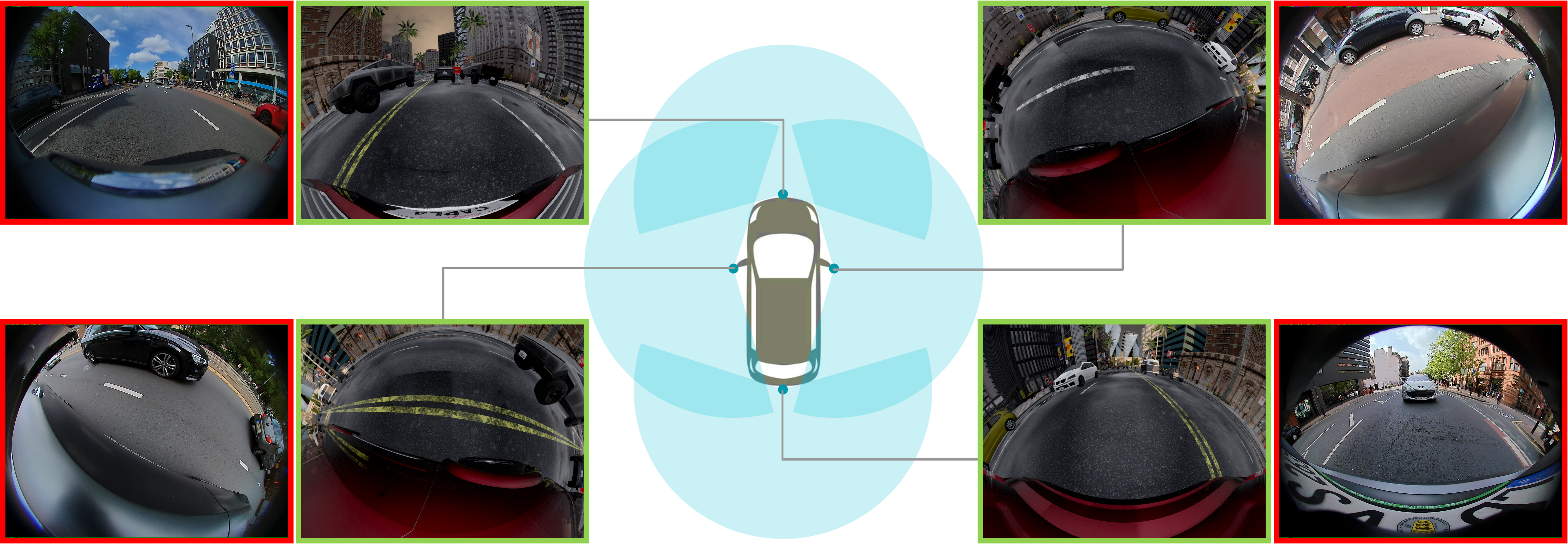} \caption{\textbf{Sample images from the surround-view camera
network showing wide field of view and $360\degree$ coverage. Real WoodScape images are marked in red  and synthetic SynWoodScape images are marked in green. }}
    \label{fig:surround-view}
\end{figure*}

Surround-view fisheye cameras have been deployed in premium cars for over ten years, starting from visualization applications on dashboard display units to provide near-field perception for automated parking. Fisheye cameras have a strong radial distortion that cannot be corrected without disadvantages, including reduced FoV and resampling distortion artifacts at the periphery~\cite{kumar2020unrectdepthnet}. Appearance variations of objects are larger due to the spatially variant distortion, particularly for close-by objects. Thus fisheye perception is a challenging task, and it is relatively less explored than pinhole cameras. 
Surround-view cameras consisting of four fisheye cameras are sufficient to cover the near-field perception as shown in Fig. \ref{fig:surround-view}. Most algorithms are usually designed to work on rectified pinhole camera images. The naive approach to operating on fisheye images is to first rectify the images and then directly apply these standard algorithms. However, such an approach carries significant drawbacks due to the reduced field-of-view and resampling distortion artifacts in the periphery of the rectified images. Furthermore, a recent comparative study \cite{Sekkat2022comparativestudy} on omnidirectional images, including fisheye showed that there is no need to rectify the fisheye images to achieve good results for semantic segmentation tasks.\par

Fisheye cameras are used in for AD tasks such as perception which involves object detection~\cite{9800124, rashedfisheyeyolo}, soiling detection~\cite{uricar2021let, das2020tiledsoilingnet}, semantic segmentation \cite{sobh2021adversarial, dahal2021roadedgenet}, weather classification \cite{dhananjaya2021weather}, depth prediction \cite{kumar2020fisheyedistancenet, kumar2021fisheyedistancenet++, kumar2020syndistnet, 9626604, kumar2020unrectdepthnet}, moving object detection \cite{yahiaoui2019fisheyemodnet} and SLAM \cite{gallagher2021hybrid, kumar2018near, kumar2018monocular} are challenging due to the highly dynamic and interactive nature of surrounding objects in the automotive scenarios~\cite{ruping2022inspect}. 
Fisheye cameras are also used commonly in other domains like video surveillance \cite{kim2016fisheye} and augmented reality \cite{schmalstieg2016augmented}. Despite its prevalence, there are only a few public datasets for fisheye images publicly available, and thus relatively little research is performed. The Oxford Robot car dataset \cite{maddern20171} is one such dataset providing fisheye camera images for AD. It contains over 100 repetitions of a consistent route through Oxford, the UK, captured over a year and used widely for long-term localization and mapping. \rd{KITTI-360 \cite{liao2022kitti} is a dataset containing fisheye and perspective images using multiple cameras including two fisheye facing each side mounted on the car roof. KITTI-360 provides ground truth annotations for several tasks but no ground truth for the fisheye images.} OmniScape \cite{sekkat2020omniscape} is a synthetic dataset providing semantic segmentation annotations and depth maps for omnidirectional cameras mounted on a motorcycle.  

\rd{
\textbf{Contributions:}
In \tablename~\ref{tab:SummaryTableFisheye}, we compare the properties of the few available automotive fisheye datasets. In particular, it can be observed that SynWoodScape provides a significantly improved feature set compared to WoodScape. In \tablename~\ref{tab:SummaryTableNoFisheye}, we compare various synthetic automotive datasets illustrating an improvement relative to the base CARLA synthetic dataset upon which SynWoodScape is built. To summarize, the contributions of this work include:}

\rd{
\begin{itemize}
    \item Creation of a new synthetic dataset consisting of 80k frames for the AD perception application; To the best of our knowledge, it is the largest fisheye dataset for the AD application.
    \item Replication of camera setup and calibration of the WoodScape dataset, thus enabling an easy combination of both datasets.
    \item Creation of ground truth for pixel-wise optical flow and depth which is not feasible to obtain densely and accurately on real scenes, Lidar cannot cover near-field regions needed for fisheye cameras.
    \item Creation of ground truth for bird's eye view tasks which takes in all four cameras as input and produces segmentation, occupancy flow, or height maps.
    \item Creation of fisheye event camera signals for evaluation of sparse event signal algorithms, and publishing the first dataset of its kind. 
    \item Experimental evaluation of domain gap between the real and synthetic fisheye datasets for various tasks.
\end{itemize}
}

\begin{table}[t]
\centering
\captionsetup{singlelinecheck=false, font=footnotesize, skip=0pt, belowskip=-12pt}
\caption{\textbf{Summary of various AD datasets containing fisheye images.} 
}
\label{tab:SummaryTableFisheye}
\resizebox{\columnwidth}{!}{
\begin{tabular}{@{}l|c|c|c|c|c|@{}}
                     & \rotatebox[origin=l]{90}{Oxford Robot Car \cite{maddern20171}} 
                     & \rotatebox[origin=l]{90}{KITTI-360 \cite{liao2022kitti}} 
                     & \rotatebox[origin=l]{90}{OmniScape \cite{sekkat2020omniscape}} 
                     & \rotatebox[origin=l]{90}{WoodScape \cite{yogamani2019woodscape}} 
                     & \rotatebox[origin=l]{90}{SynWoodScape (proposed)~} \\ \midrule
Real/Synthetic       & Real          & Real          & Synthetic     & Real          & Synthetic     \\ \midrule
Ego Vehicle          & Car           & Car           & Motorcycle    & Car           & Car           \\ \midrule
Fisheye Resolution   & 1024×1024     & 1400×1400     & 1024×1024     & 1280×966      & 1280×966      \\ \midrule
Fisheye HFoV         & $180^\degree$ & $180^\degree$ & $185^\degree$ & $190^\degree$ & $190^\degree$ \\ \midrule
Bird's Eye View      & \xm           & \xm           & \xm           & \xm           & \ch           \\ \midrule
Semantic Seg.        & \xm           & \xm           & \ch           & \ch           & \ch           \\ \midrule
Instance Seg.        & \xm           & \xm           & \ch           & \ch           & \ch           \\ \midrule
Motion Seg.          & \xm           & \xm           & \xm           & \ch           & \ch           \\ \midrule
2D/3D Bounding Boxes & \xm           & \ch           & \xm           & \ch           & \ch           \\ \midrule
Depth Map            & \xm           & \xm           & \ch           & \xm           & \ch           \\ \midrule
Event Camera Signals & \xm           & \xm           & \xm           & \xm           & \ch           \\ \midrule
Optical Flow         & \xm           & \xm           & \xm           & \xm           & \ch           \\ \midrule
Lidar                & \ch           & \ch           & \ch           & \ch           & \ch           \\ \midrule
IMU                  & \ch           & \ch           & \ch           & \ch           & \ch           \\ \midrule
GNSS                 & \ch           & \ch           & \ch           & \ch           & \ch           \\ \midrule
\end{tabular}}
\end{table}

\section{SynWoodScape}

The SynWoodScape dataset is a synthetic version of the WoodScape dataset. The same configuration used to acquire the real data from different locations in Europe and the USA is used in CARLA Simulator (release 0.9.10.1). The same calibration parameters, intrinsic and extrinsic ones, were also used to simulate the different sensors. The use of the simulator allows us to extract, in addition to all the ground truths proposed in the WoodScape dataset, the ground truths for pixel-wise tasks like depth map, optical flow, and event camera signals in a very precise manner. It also allows us to extract time synchronized images from four fisheye surround-view cameras in addition to a bird's eye view (BEV) image. We also used the simulator to extract images in different weather and lighting conditions. In the following subsections, we explain the construction of the fisheye images using the calibration parameters of the WoodScape dataset and the computation of the different ground truths.
\subsection{Fisheye image generation}

\begin{figure*}[tb]
    \centering
    \captionsetup{singlelinecheck=false, font=footnotesize, belowskip=-12pt}
    \includegraphics[width=0.9\textwidth]{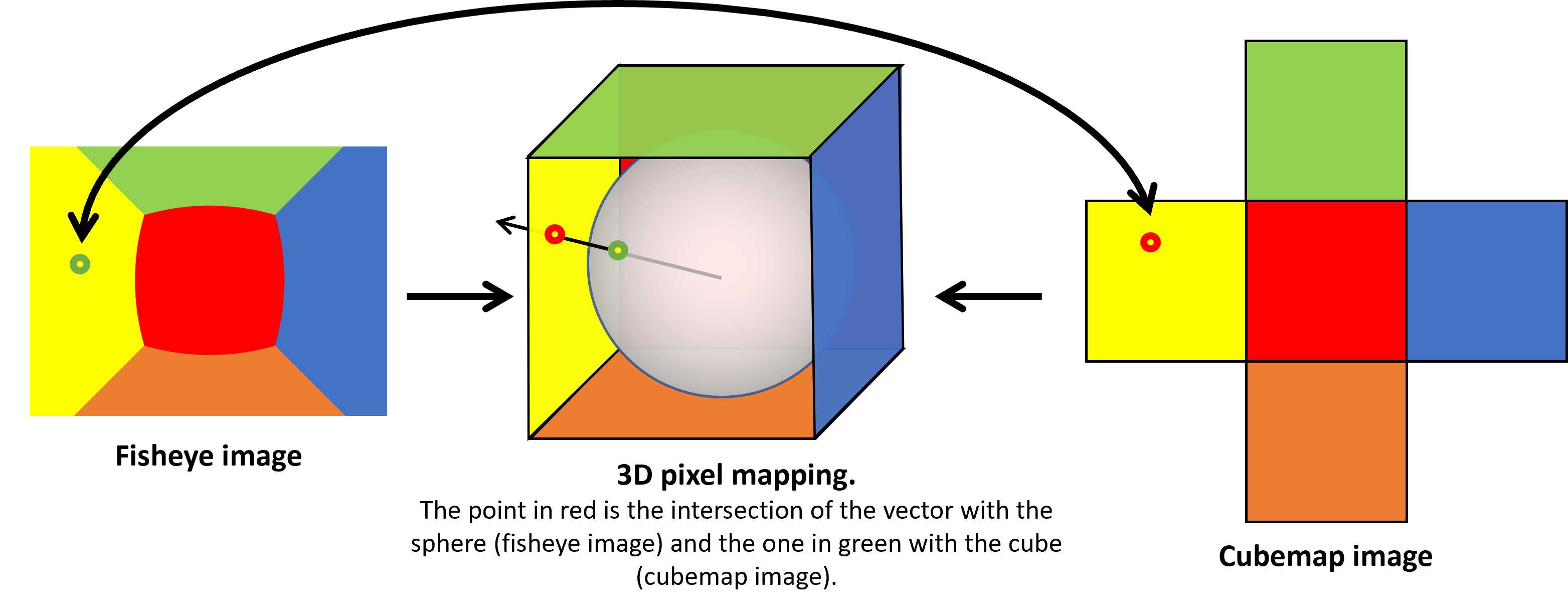}
    \caption{\textbf{Mapping of the cubemap image's pixels to the fisheye image.}}
    \label{fig:fisheye-projection}
\end{figure*}

To generate the fisheye images, we used a framework based on the cubemap representation of a $360^\degree$ image and the calibration model proposed in the WoodScape dataset \cite{yogamani2019woodscape}. The model uses a fourth-order polynomial function to estimate the mapping of incident angle to image radius in pixels $(r(\theta)= a_1\theta + a_2\theta^2 + a_3\theta^3 + a_4\theta^4)$. Using this model, each pixel in the fisheye image can be associated with a 3D direction on the unit sphere. We also construct a unit cube that corresponds to the cubemap image. Using ray tracing from the center of the sphere and the cube, we compute the pixel mapping between the fisheye and the cubemap images, as sketched in Fig. \ref{fig:fisheye-projection}. The mapping of the cubemap image to the fisheye image is obtained by the intersection of the 3D direction with both the sphere and the cube. A lookup table is then built for each fisheye camera to store the correspondences between the two representations. To extract the fisheye images from CARLA, we acquired five images that form the five views of the cubemap needed to build the fisheye image, and we used the exact calibration parameters of the cameras used to acquire the WoodScape dataset \cite{yogamani2019woodscape} to build the sphere and to place the cameras using the same positions and rotations relative to the car. In such a way, we preserve the same configuration of the WoodScape dataset as if we used the same acquisition platform inside CARLA Simulator.\par

\begin{table}[t]
  \captionsetup{singlelinecheck=false, font=footnotesize, belowskip=-12pt}
\centering
\caption{\textbf{Summary of various AD synthetic datasets.}}
\label{tab:SummaryTableNoFisheye}
\resizebox{\columnwidth}{!}{
\begin{tabular}{@{}l|c|c|c|c|c|c|@{}}
                      & \rotatebox[origin=l]{90}{SYNTHIA \cite{SYNTHIA}} 
                      & \rotatebox[origin=l]{90}{Driving in the Matrix \cite{johnson2017driving}}
                      & \rotatebox[origin=l]{90}{Playing for benchmarks \cite{Richter_2017}}   
                      & \rotatebox[origin=l]{90}{Apollo Synthetic \cite{wang2019apolloscape}} 
                      & \rotatebox[origin=l]{90}{All-in-One Drive \cite{Weng2020_AIODrive}}
                      & \rotatebox[origin=l]{90}{SynWoodScape (proposed)} \\ \midrule
Semantic Seg.         & \ch               & \ch   & \ch   & \ch    & \ch    & \ch       \\ \midrule
Instance Seg.         & \xm               & \xm   & \ch   & \ch    & \ch    & \ch       \\ \midrule
Motion Seg.           & \xm               & \xm   & \xm   & \xm    & \xm    & \ch       \\ \midrule
2D Bounding Boxes     & \xm               & \ch   & \ch   & \ch    & \ch    & \ch       \\ \midrule
3D Bounding Boxes     & \xm               & \xm   & \ch   & \ch    & \ch    & \ch       \\ \midrule
Depth Map             & \ch               & \xm   & \xm   & \ch    & \ch    & \ch       \\ \midrule
Event Camera signals  & \xm               & \xm   & \xm   & \xm    & \xm    & \ch       \\ \midrule
Optical Flow          & \xm               & \xm   & \ch   & \xm    & \xm    & \ch       \\ \midrule
Lidar                 & \xm               & \xm   & \xm   & \xm    & \ch    & \ch       \\ \midrule
Semantic Lidar        & \xm               & \xm   & \xm   & \xm    & \ch    & \ch       \\ \midrule
Radar                 & \xm               & \xm   & \xm   & \xm    & \ch    & \ch       \\ \midrule
IMU                   & \xm               & \xm   & \xm   & \xm    & \ch    & \ch       \\ \midrule
GNSS                  & \xm               & \xm   & \xm   & \xm    & \ch    & \ch       \\ \midrule
Bird's Eye View       & \xm               & \xm   & \xm   & \xm    & \xm    & \ch       \\ \midrule
$360\degree$ Coverage & \ch               & \xm   & \xm   & \xm    & \ch    & \ch       \\ \midrule
Omnidirectional       & \ch               & \xm   & \xm   & \xm    & \xm    & \ch       \\ 
Images                & (Equirectangular) &       &       &        &        & (Fisheye) \\ \midrule
Simulator             & SYNTHIA           & GTA V & GTA V & Apollo & CARLA  & CARLA     \\ \midrule
Engine                & UNITY             & RAGE  & RAGE  & UNITY  & Unreal & Unreal    \\ \midrule
\end{tabular}}
\end{table}

\begin{figure}
  \captionsetup{singlelinecheck=false, font=footnotesize, belowskip=-10pt}
  \centering
    \includegraphics[width=\linewidth]{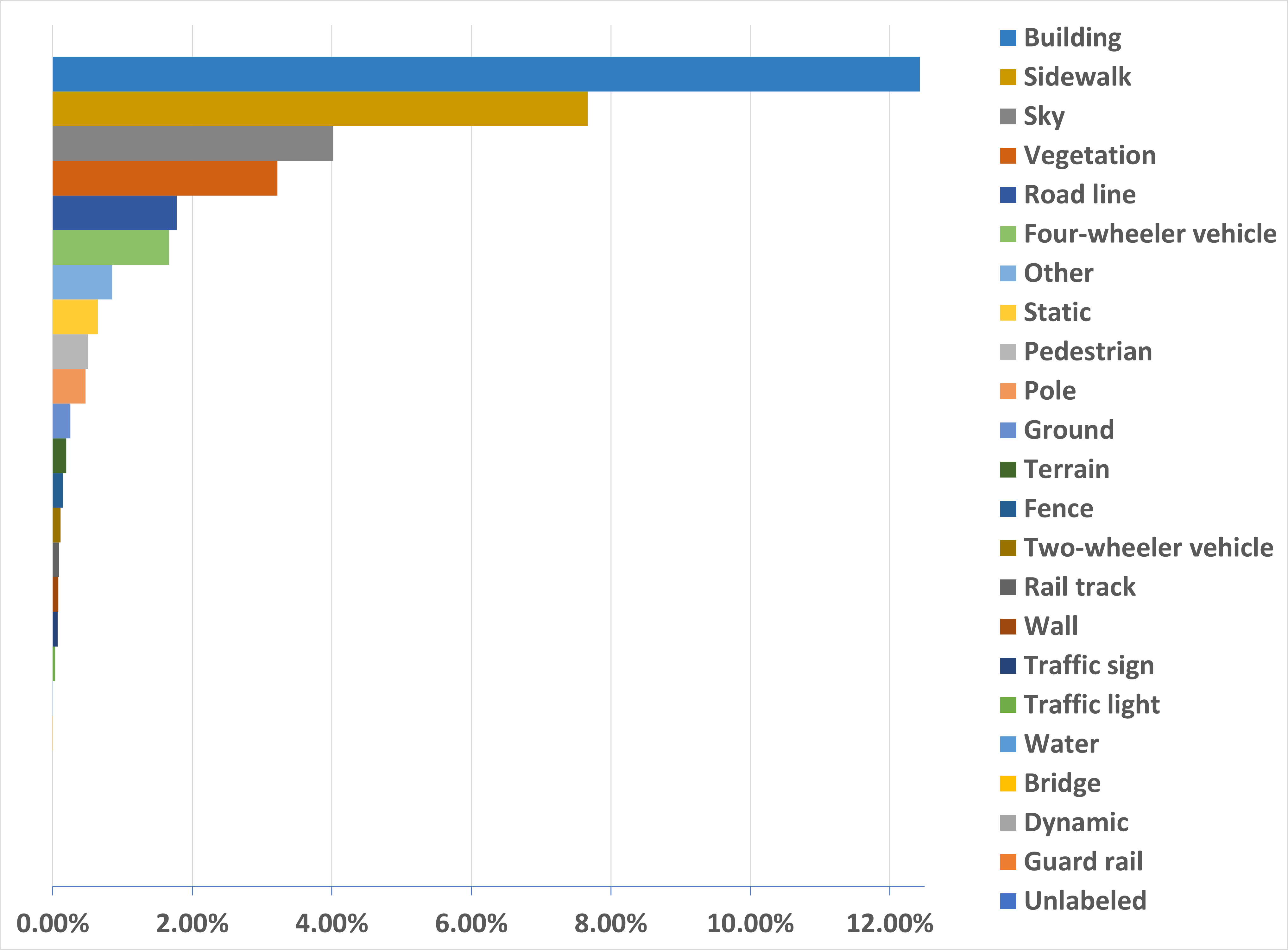} \caption{\textbf{Percentage of pixels representing all classes in the semantic segmentation ground truth of all images in the dataset including BEV images. Largest classes namely Road 42,32\% and Ego vehicle 23,45\% are not plotted due to its large size.}}
    \label{fig:pixels_percentages}
\end{figure}

\subsection{Dataset Details}

The SynWoodScape dataset contains synthetic data generated from CARLA Simulator \cite{carlasimulator}, each sample out of the 10k samples provided contains surround-view fisheye images in addition to bird's eye view and front view perspective images. Each image comes with a previous image (for tasks that require two consecutive frames) and ground truth for multiple tasks namely, semantic segmentation into 25 classes, instance segmentation, motion segmentation, depth map, optical flow, event camera signals, 3D and 2D bounding boxes, lidar data, radar dara, IMU and GNSS data. \rd{\figurename~\ref{One_capture} lists all the images with the corresponding ground truth images from a single sample; the current and the previous RGB images are merged to better show the movements of objects in the scene.} The acquisition was made using a frame rate of 10 FPS. The intrinsic and extrinsic parameters of the used cameras are similar to the parameters of the real cameras used in the acquisition of the WoodScape dataset \cite{yogamani2019woodscape}. The dimensions of the fisheye images are $1280\times966$, of the bird's eye view images are $1024\times1024$ and of the front view images are $3264\times2448$. \figurename~\ref{fig:teaser} and \figurename~\ref{One_capture} shows an example of images extracted with generated ground truth data.

\rd{Various random scenarios are present in the dataset. Just urban scene images are considered using the HD Town10, which is a city with different environments such as an avenue and promenade. The used synthetic environment is around 250 meters by 300 meters, and there are 155 recommended spawn points for vehicles. Each time a random spawn point is chosen to spawn the ego vehicle using a random color, the rest of the spawn points are used to spawn other random vehicles from a set of vehicles. Regarding pedestrians; the maximum possible candidates are spawned around the ego vehicle, their amount varies depending on the possible positions to spawn a pedestrian and also on the available computation resources. It results in a minimum of 125 vehicles and pedestrians present on the scene and a maximum of 289. All the vehicles including the ego vehicle and also the pedestrians are controlled automatically using the Traffic Manager provided by Carla Simulator, which manages the urban traffic using an autopilot mode to simulate natural behaviors. The images are captured in nine different weather and lightning conditions predefined in the simulator: Clear Noon, Clear Sunset, Cloudy Noon, Cloudy Sunset, Default, Wet Cloudy Noon, Wet Cloudy Sunset, Wet Noon, Wet Sunset.}

\begin{figure}[t]
  \captionsetup{singlelinecheck=false, font=footnotesize, belowskip=-12pt}
  \centering
    \includegraphics[width=\linewidth]{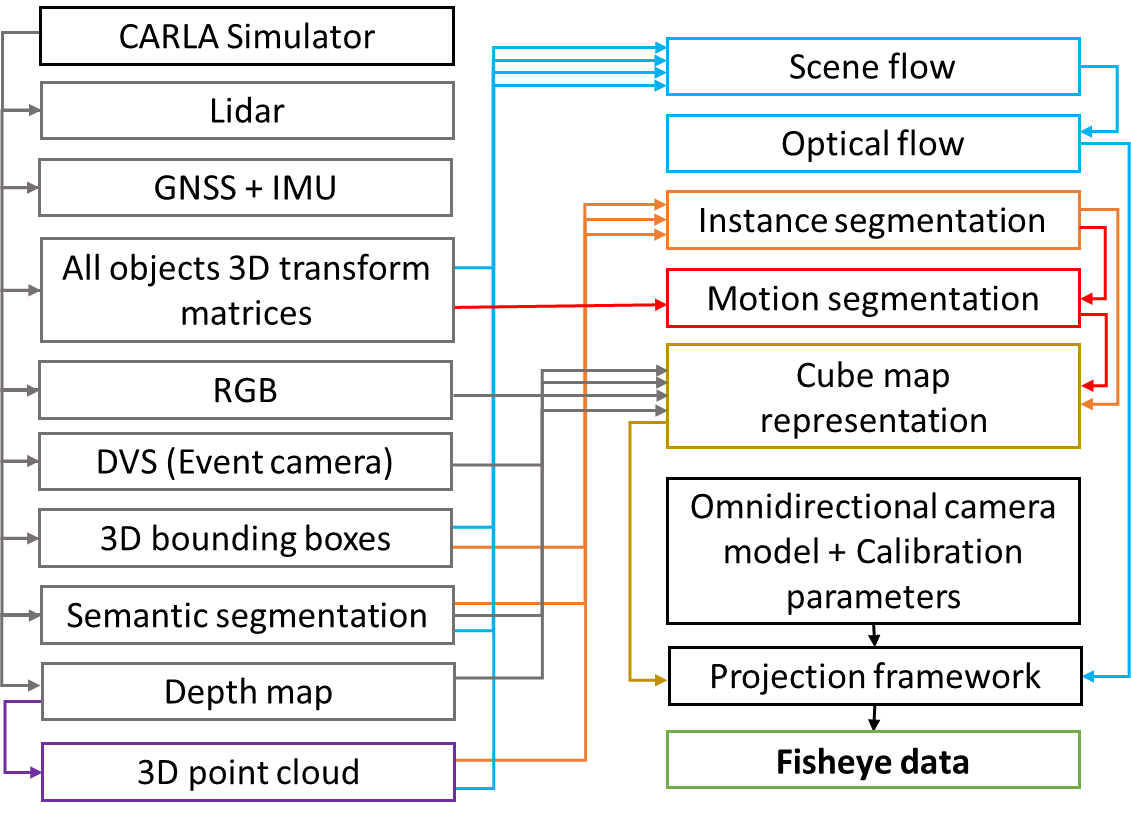} \caption{\textbf{Illustration of data extraction procedure from CARLA Simulator.}}
    \label{fig:SynWoodScape_diagram}
\end{figure}

\rd{In the SynWoodScape, the 2D/3D bounding boxes include four-wheeler vehicles, two-wheeled vehicles, and pedestrians. In \tablename~\ref{tab:BBoxes_motion_stat} object statistics are made showing the frequencies of each class across all frames, as well as motion segmentation statistics using different thresholds. The semantic segmentation is provided into the following classes: unlabeled, building, fence, other, pedestrian, pole, road line, road, sidewalk, vegetation, four-wheeler vehicle, wall, traffic sign, sky, ground, bridge, rail track, guard rail, traffic light, water, terrain, two-wheeler vehicle, static, dynamic, ego vehicle. \figurename~\ref{fig:pixels_percentages} shows the distribution of pixels of all images in the dataset.}
\rd{\figurename~\ref{fig:surround-view} shows side by side surround-view fisheye images from the WoodScape and the SynWoodScape dataset. 
\figurename~\ref{fig:SynWoodScape_diagram} shows a simplified diagram of the extraction procedure of all ground truth data from the CARLA Simulator. In the following section, we explain how we compute the ground truths that are not directly extracted from CARLA Simulator. It is worth noting that the following methods can be used also for other simulators or datasets, as long as the same inputs are available.}

\begin{table}
  \captionsetup{font=footnotesize, belowskip=-2pt}
\centering
\caption{\textbf{
Statistics of objects in the dataset. The second grouped column shows the frequency of all objects. The third grouped column shows statistics of moving objects thresholded according to distance traveled across consecutive frames.
}}
\resizebox{\columnwidth}{!}{
\begin{tabular}{@{}l|cc|ccccc@{}}
\toprule
\textbf{\textit{Class}}       & \multicolumn{2}{c|}{\textbf{\textit{All objects}}}  & \multicolumn{5}{c}{\textbf{\textit{Moving objects}}} \\
                     & \multicolumn{2}{c|}{\textit{Frequency}}  & \multicolumn{5}{c}{\textit{Thresholds in meters}} \\ 
\multicolumn{1}{l|}{} & \% of images & objects/image & 0.0      & 0.25     & 0.5      & 0.75    & 1.0    \\
\midrule
Pedestrian           & 98.68               & 34.09         & 4.76     & 3.15     & 1.15     & 0.35    & 0.0    \\
Four-wheeler  & 90.44               & 8.24          & 0.68     & 0.45     & 0.13     & 0.04    & 0.0    \\
Two-wheeler   & 80.72               & 2.89          & 0.23     & 0.16     & 0.06     & 0.02    & 0.0    \\ 
\bottomrule
\end{tabular}}
\label{tab:BBoxes_motion_stat}
\end{table}

\subsection{Instance Segmentation}
\label{sec:Instance Segmentation}

To extract the instance segmentation, we used the depth maps, the 3D bounding boxes, and the semantic segmentation ground truth. With these three modalities, we developed a tool to compute the instance segmentation on perspective images used to generate the omnidirectional images. This tool is based on ray tracing. For each pixel, we compute the 3D position in the world reference of the CARLA Simulator. This is achieved by using the depth map and the camera transform matrix from the sensor to the world reference, which can be obtained after computing the focal length of the camera. The camera transform matrix is obtained according to 
\begin{equation}
   \label{formula1}
   K=\left(\begin{array}{lll}
{f} & {0} & {w / 2} \\
{0} & {f} & {h/2} \\
{0} & {0} & {1}
\end{array}\right),
\end{equation}
where $w$ and $h$ are the width and the height of the image, respectively, and $f$ is the focal length of the camera and it is computed using the formula:
\begin{equation}
   \label{formula2}
f=\frac{w}{2 \tan \left(\frac{\pi  fov}{360}\right)},
\end{equation}
where $fov$ the field of view of the camera.
The 3D position of the pixel $p$ of coordinate $(x,y)$ is obtained using the following formula, where $d$ is the corresponding depth map value:
\begin{equation}
   \label{formula3}
\left(\begin{array}{l}
{X} \\
{Y} \\
{Z}
\end{array}\right)=K^{-1}\left(\begin{array}{l}
{x} \\
{y} \\
{1}
\end{array}\right) d\cdot
\end{equation}
After computing the 3D points of all pixels using \eqref{formula3}, and since we have the 3D bounding boxes of each object in the scene identifiable by a unique id, we need to check which of these 3D bounding boxes the computed 3D points are inside. We check this by computing the six planes formed by the bounding box; if the 3D points are in between the parallel planes, the point is considered inside the bounding box.

\begin{figure*}[t!]
\captionsetup[subfigure]{labelformat=empty}
\centering


\begin{subfigure}{\sizeFish\textwidth}
    \includegraphics[width=\textwidth]{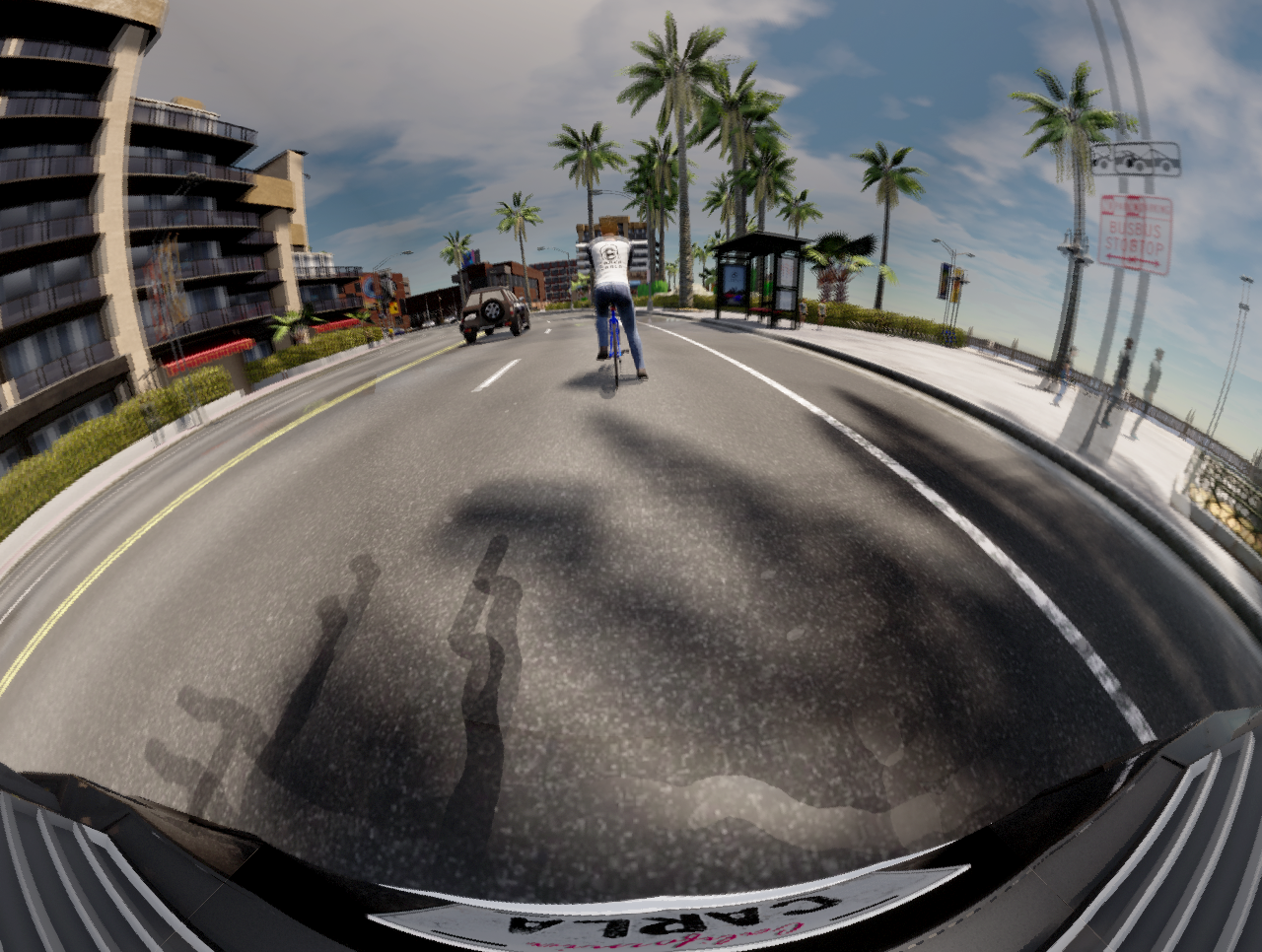}
    \vspace{-1cm}
    \caption{\textcolor{white}{(a)}}
\end{subfigure}%
\hfill
\begin{subfigure}{\sizeFish\textwidth}
    \includegraphics[width=\textwidth]{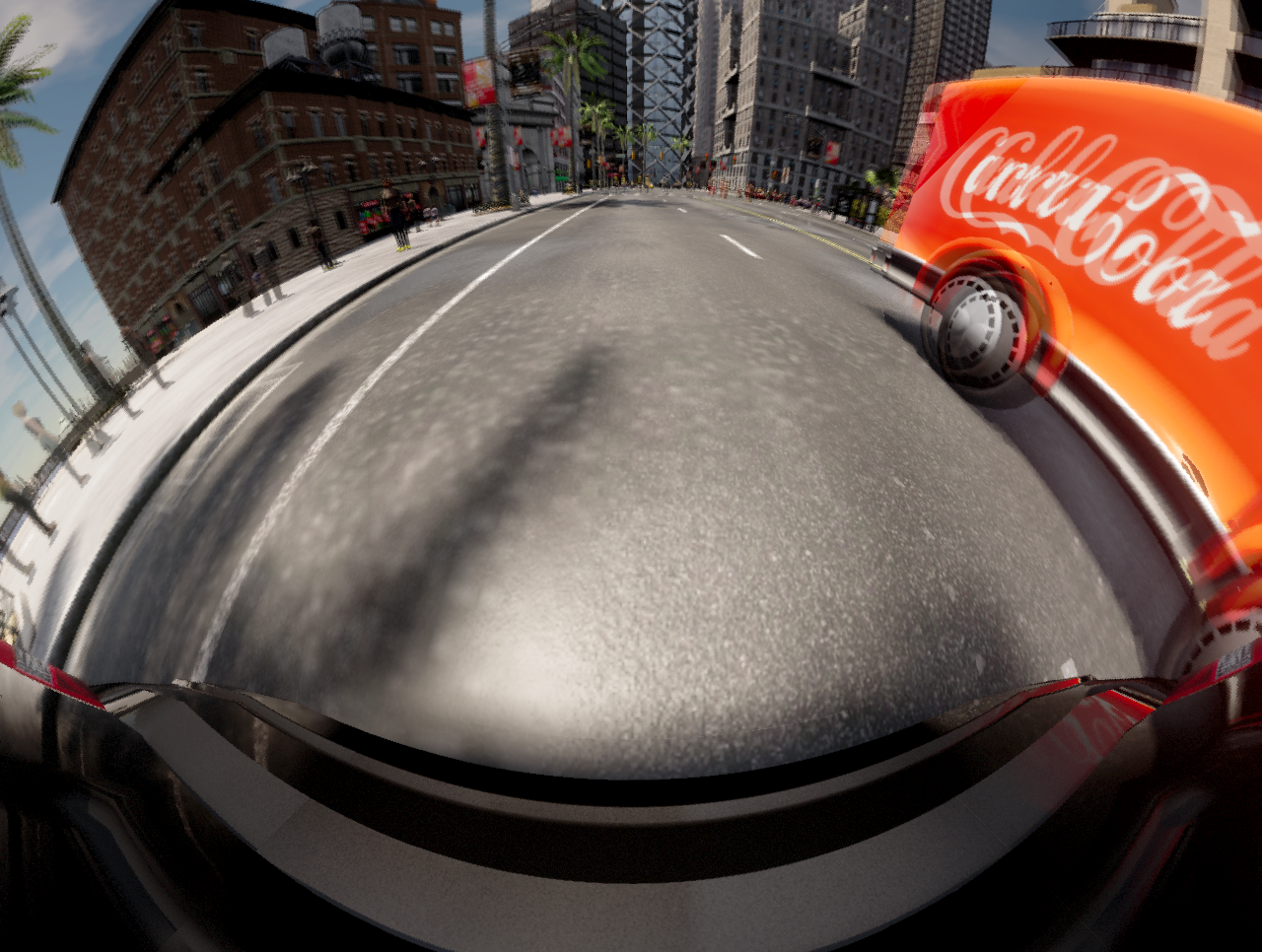}
    \vspace{-1cm}
    \caption{\textcolor{white}{(b)}}
\end{subfigure}%
\hfill
\begin{subfigure}{\sizeBEV\textwidth}
    \includegraphics[width=\textwidth]{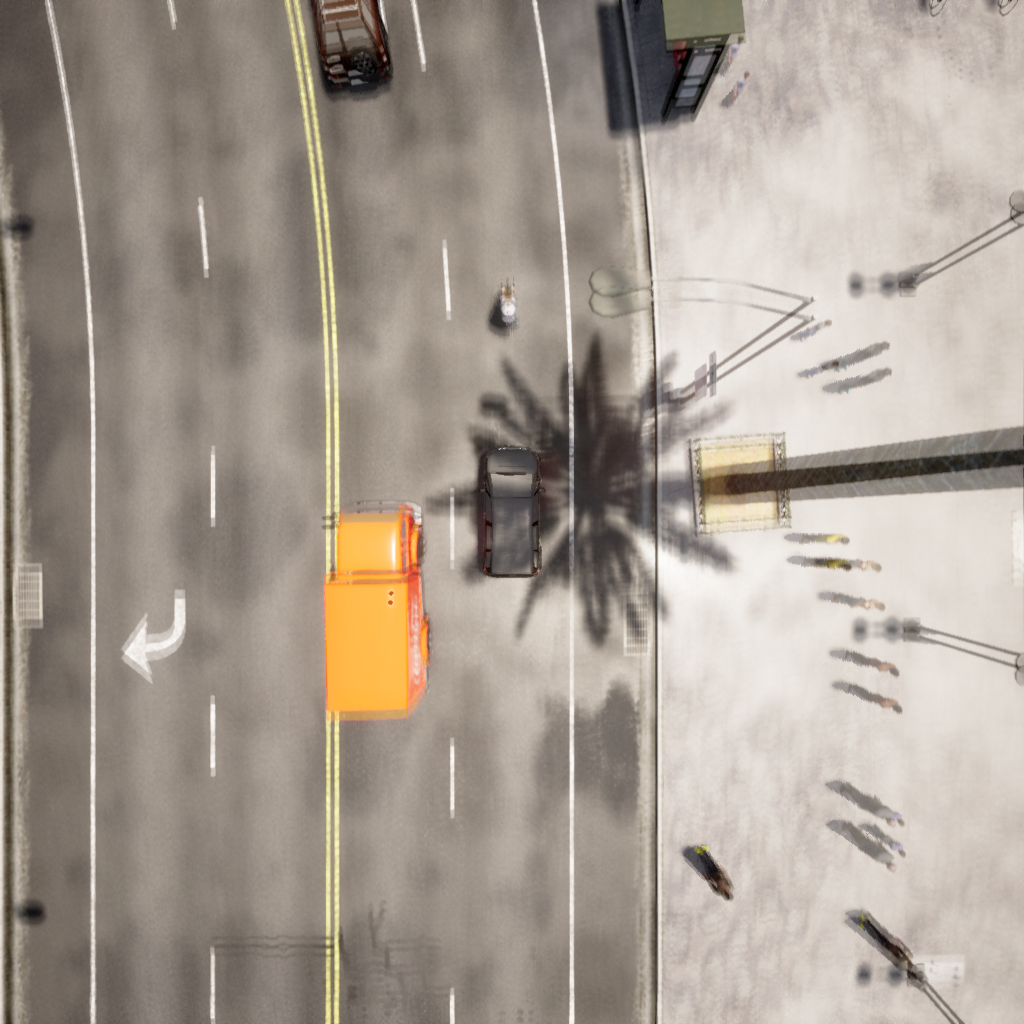}
    \vspace{-1cm}
    \caption{\textcolor{white}{(c)}}
\end{subfigure}%
\hfill
\begin{subfigure}{\sizeFish\textwidth}
    \includegraphics[width=\textwidth]{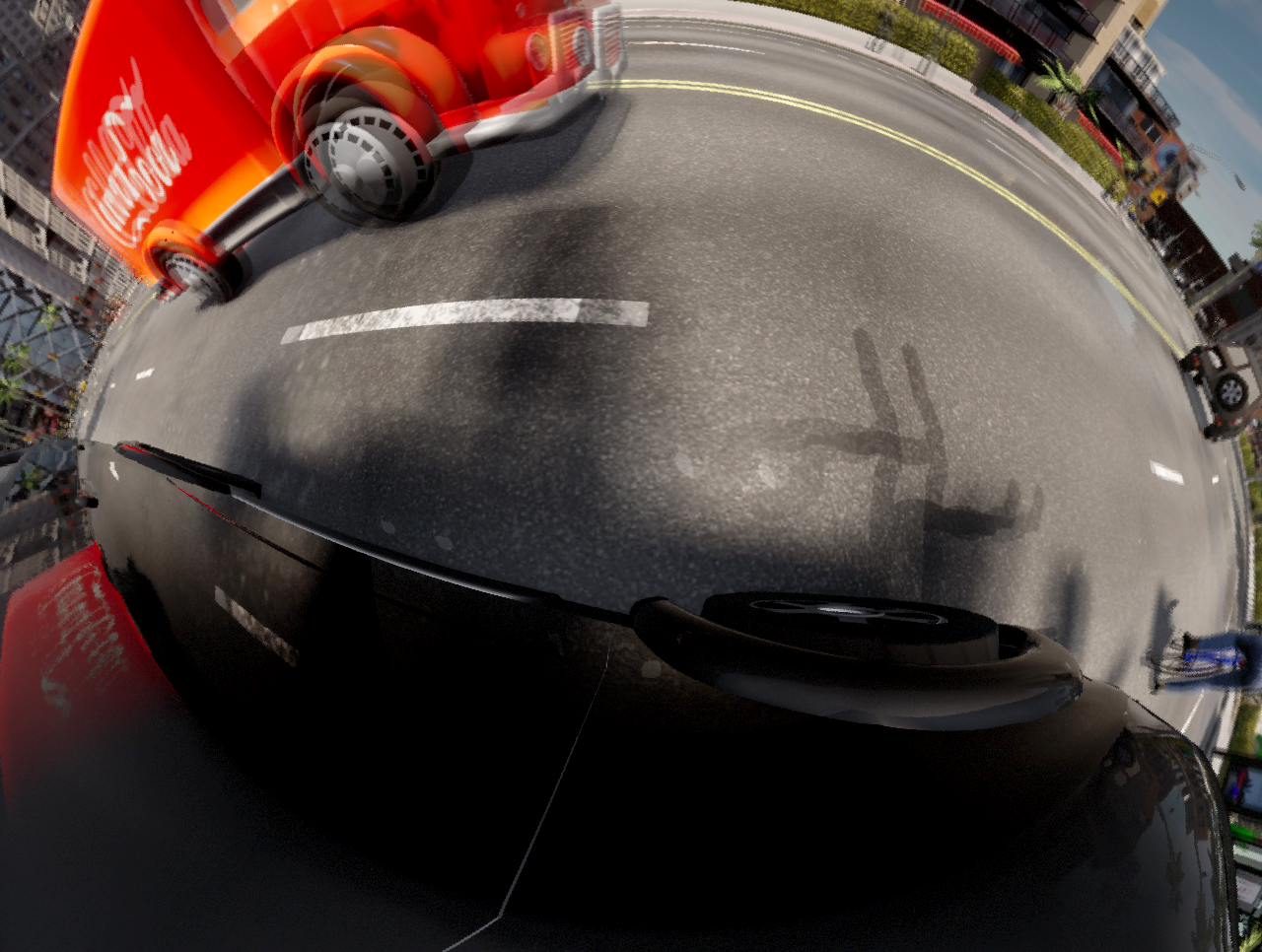}
    \vspace{-1cm}
    \caption{\textcolor{white}{(d)}}
\end{subfigure}%
\hfill
\begin{subfigure}{\sizeFish\textwidth}
    \includegraphics[width=\textwidth]{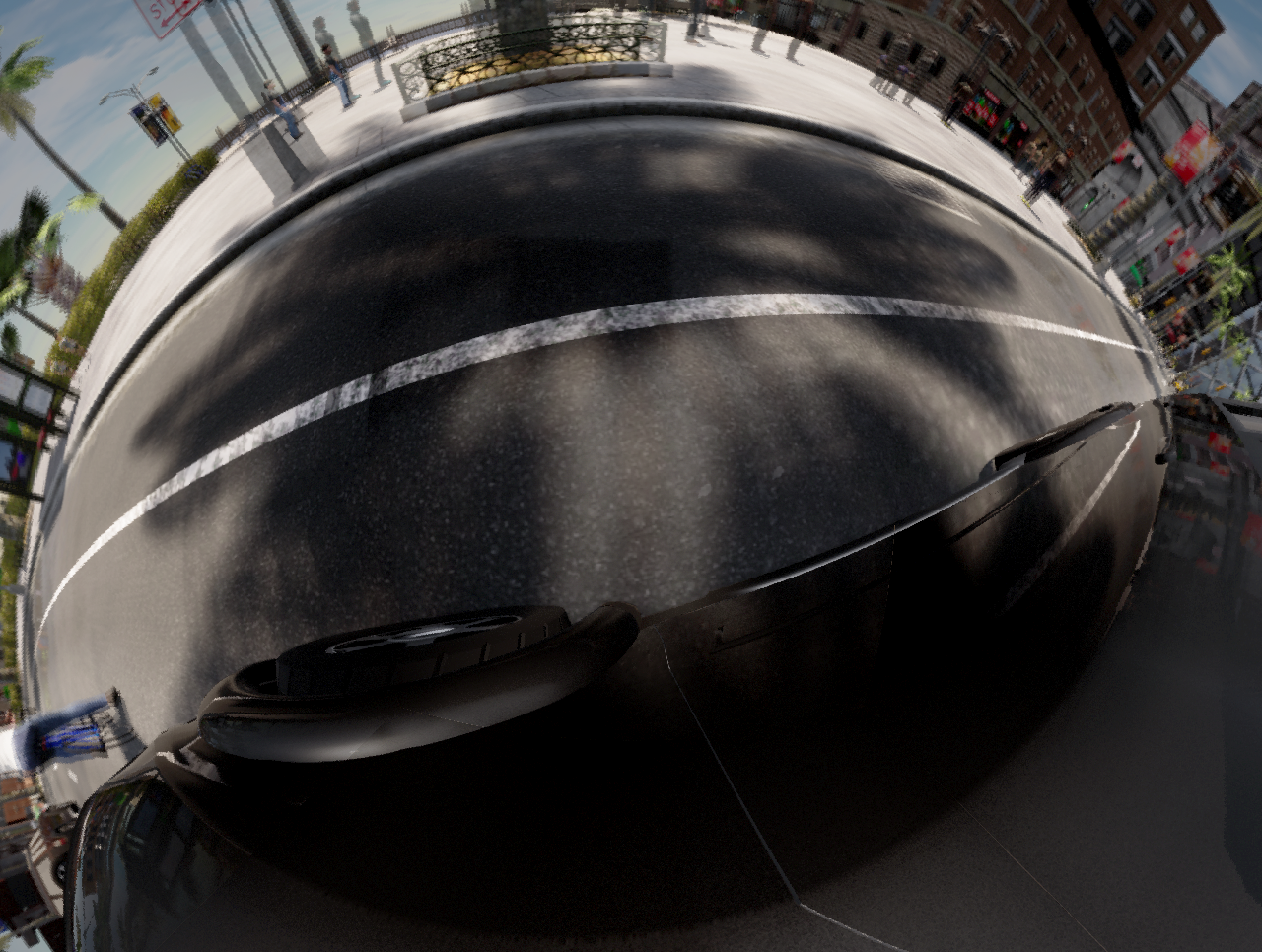}
    \vspace{-1cm}
    \caption{\textcolor{white}{(e)}}
\end{subfigure}%
\hfill
\vspace{0.07cm}


\begin{subfigure}{\sizeFish\textwidth}
    \includegraphics[width=\textwidth]{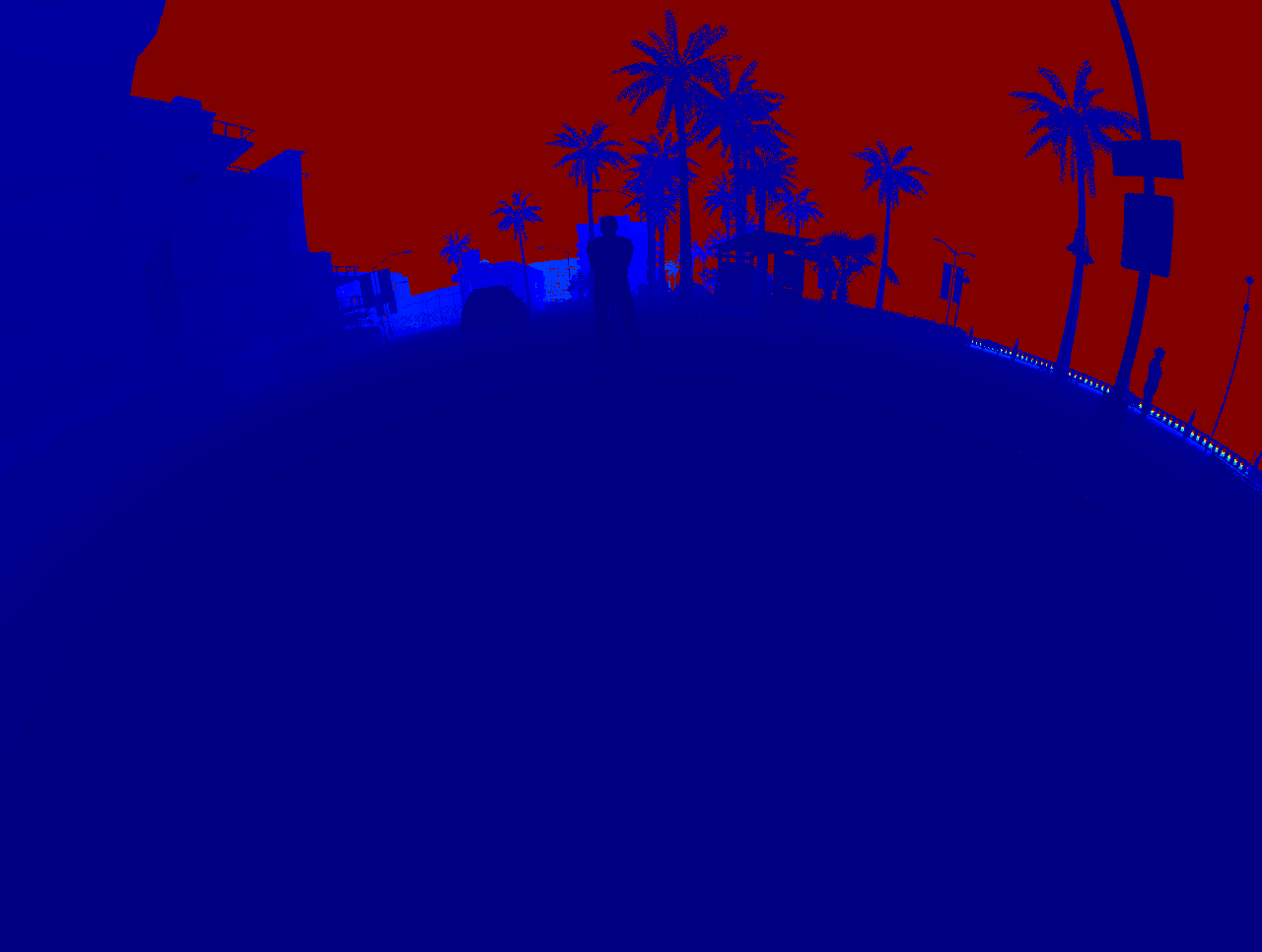}
    \vspace{-1cm}
    \caption{\textcolor{white}{(a)}}
\end{subfigure}%
\hfill
\begin{subfigure}{\sizeFish\textwidth}
    \includegraphics[width=\textwidth]{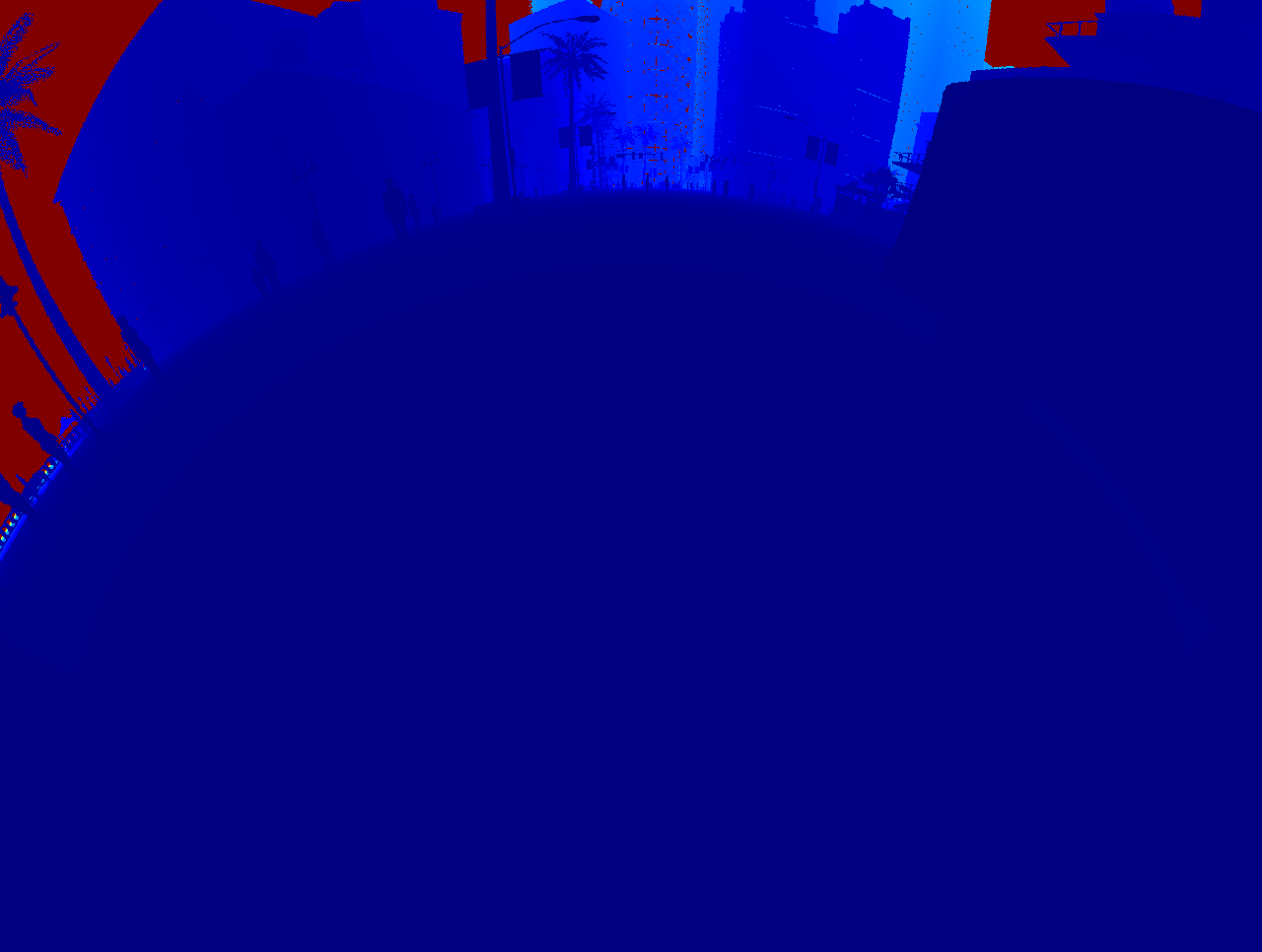}
    \vspace{-1cm}
    \caption{\textcolor{white}{(b)}}
\end{subfigure}%
\hfill
\begin{subfigure}{\sizeBEV\textwidth}
    \includegraphics[width=\textwidth]{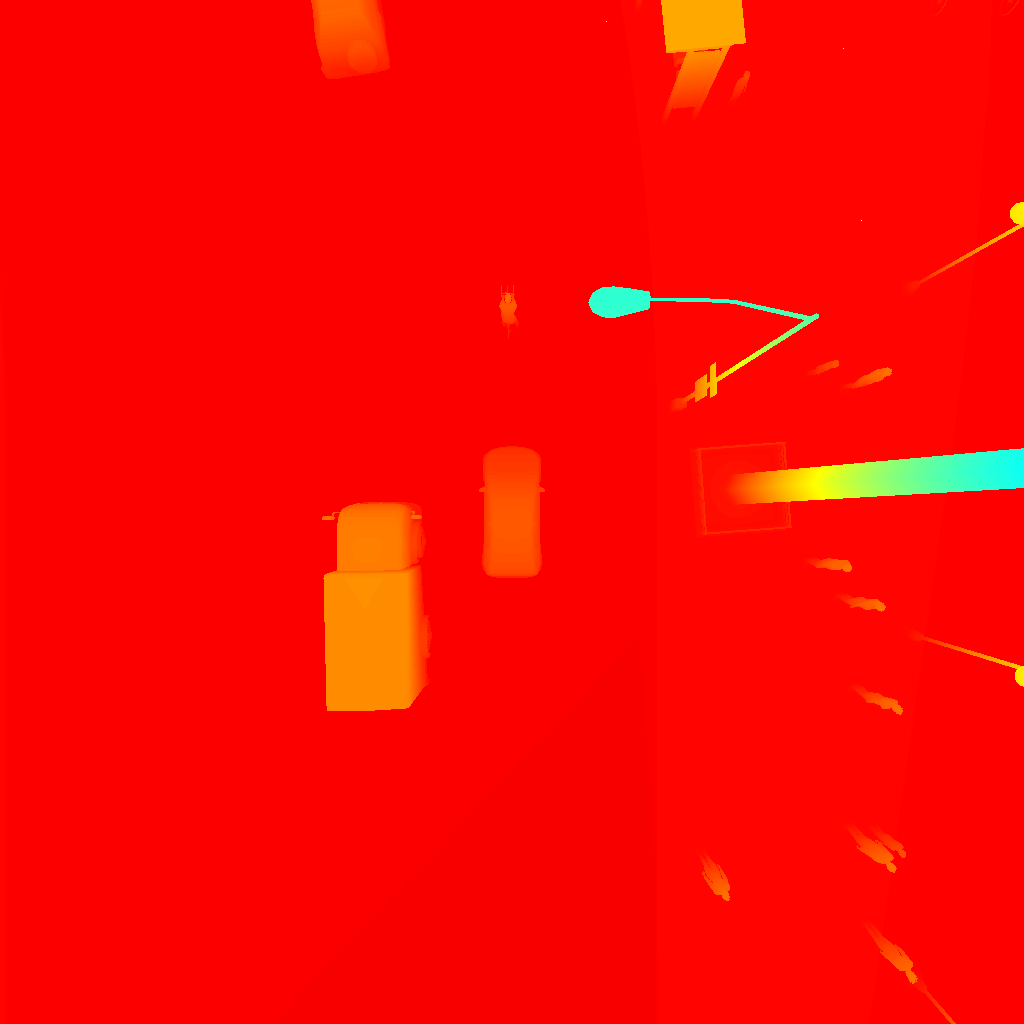}
    \vspace{-1cm}
    \caption{\textcolor{white}{(c)}}
\end{subfigure}%
\hfill
\begin{subfigure}{\sizeFish\textwidth}
    \includegraphics[width=\textwidth]{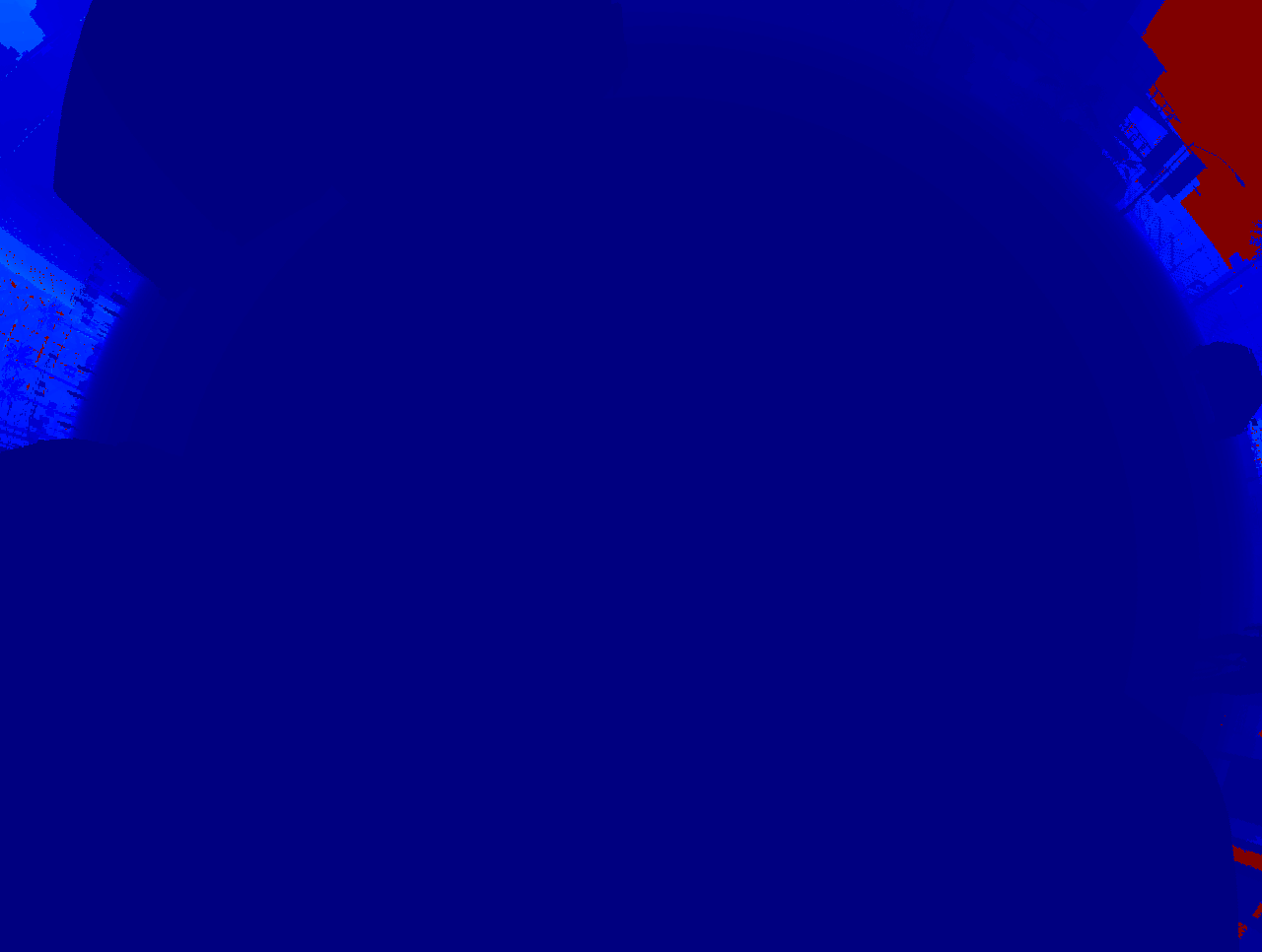}
    \vspace{-1cm}
    \caption{\textcolor{white}{(d)}}
\end{subfigure}%
\hfill
\begin{subfigure}{\sizeFish\textwidth}
    \includegraphics[width=\textwidth]{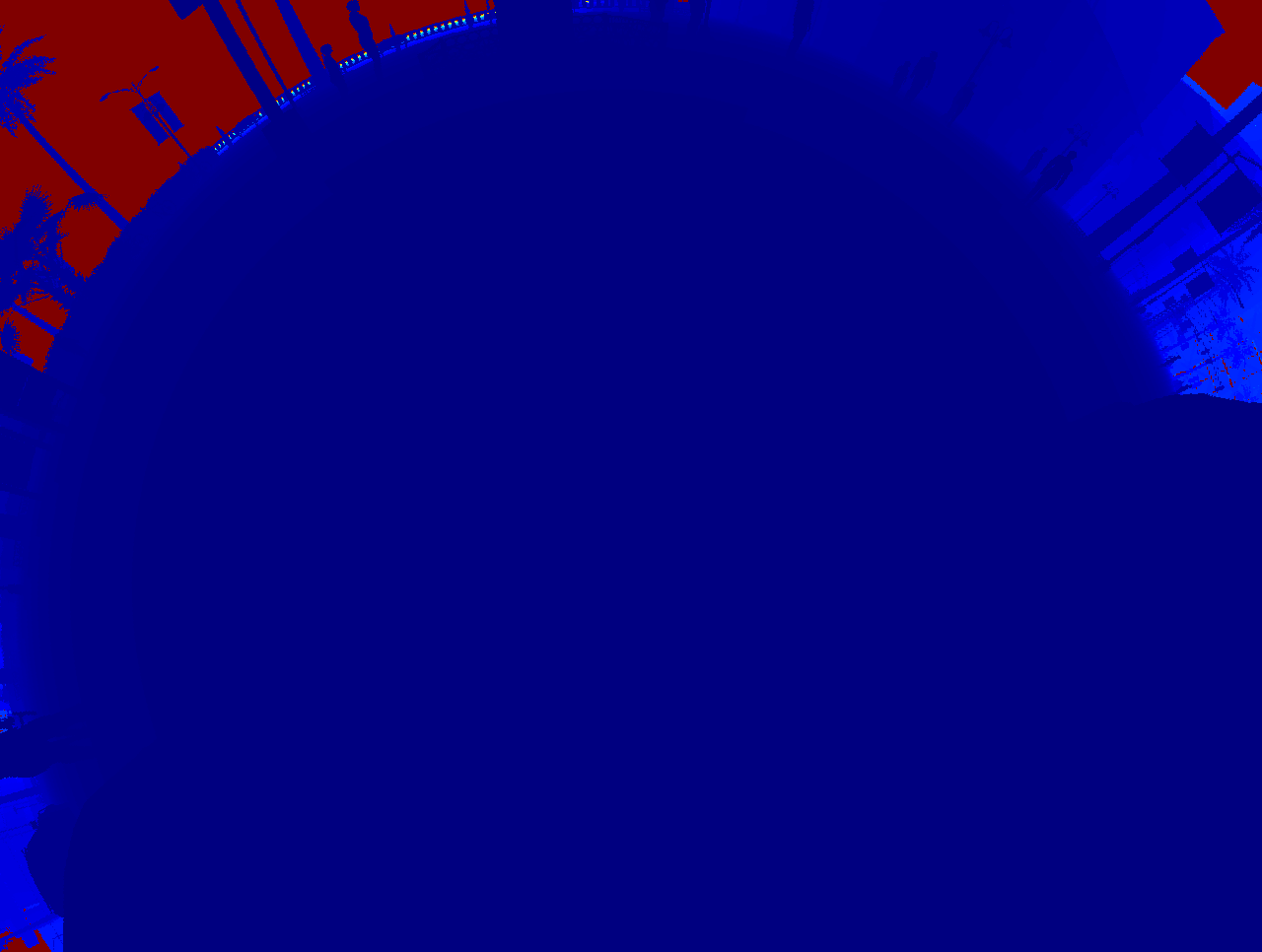}
    \vspace{-1cm}
    \caption{\textcolor{white}{(e)}}
\end{subfigure}%
\hfill
\vspace{0.07cm}


\begin{subfigure}{\sizeFish\textwidth}
    \includegraphics[width=\textwidth]{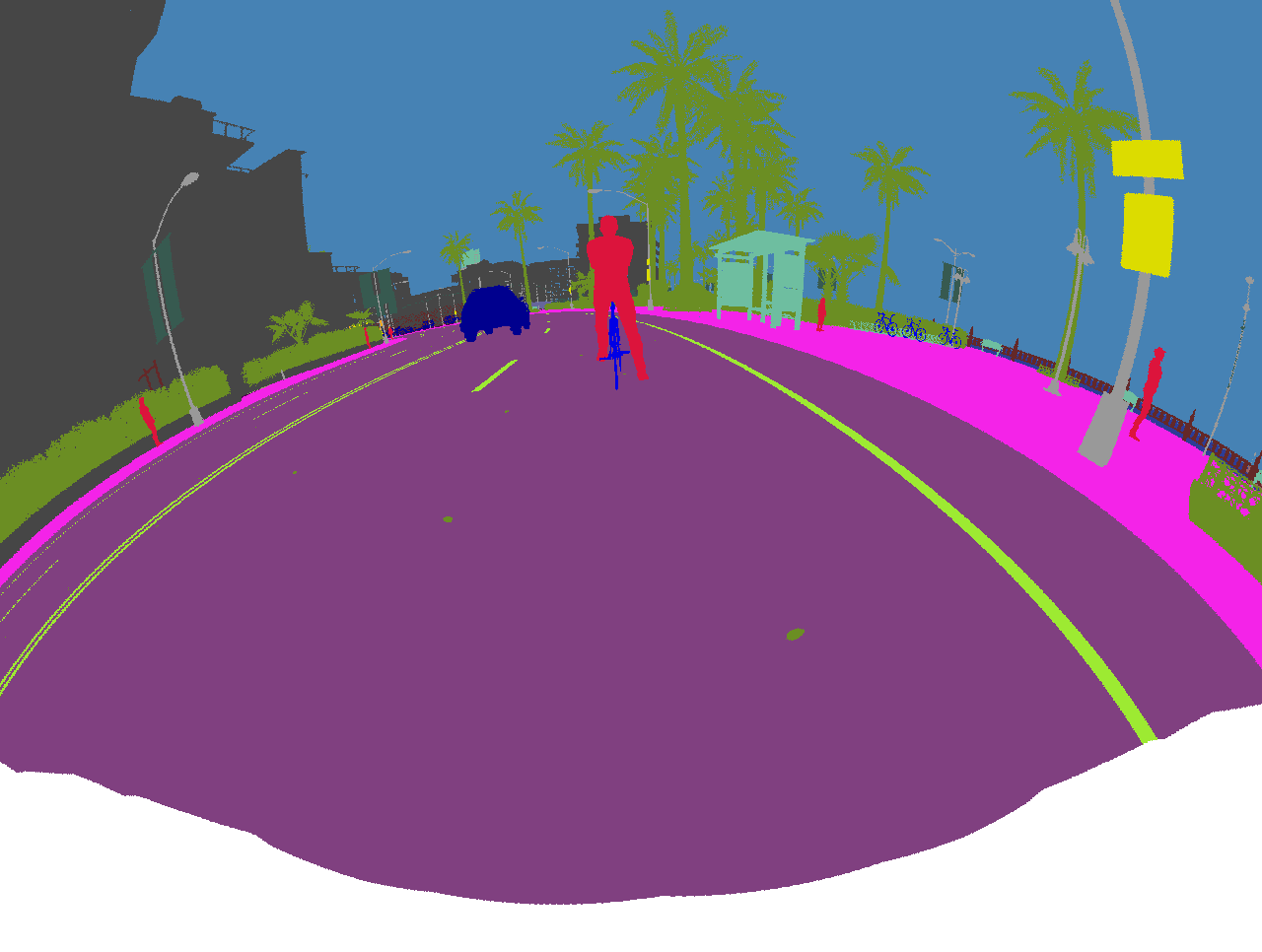}
    \vspace{-1cm}
    \caption{\textcolor{black}{(a)}}
\end{subfigure}%
\hfill
\begin{subfigure}{\sizeFish\textwidth}
    \includegraphics[width=\textwidth]{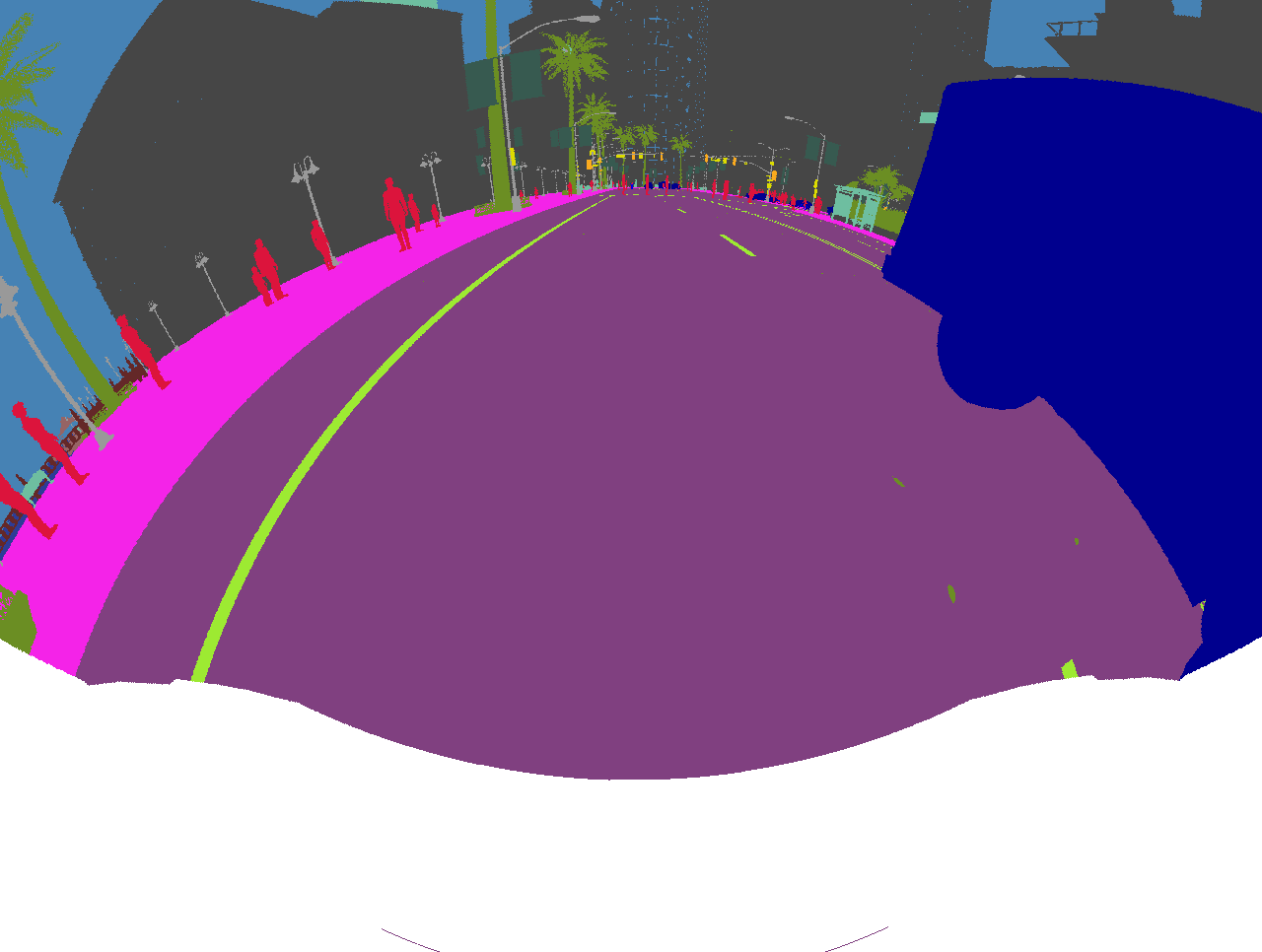}
    \vspace{-1cm}
    \caption{\textcolor{black}{(b)}}
\end{subfigure}%
\hfill
\begin{subfigure}{\sizeBEV\textwidth}
    \includegraphics[width=\textwidth]{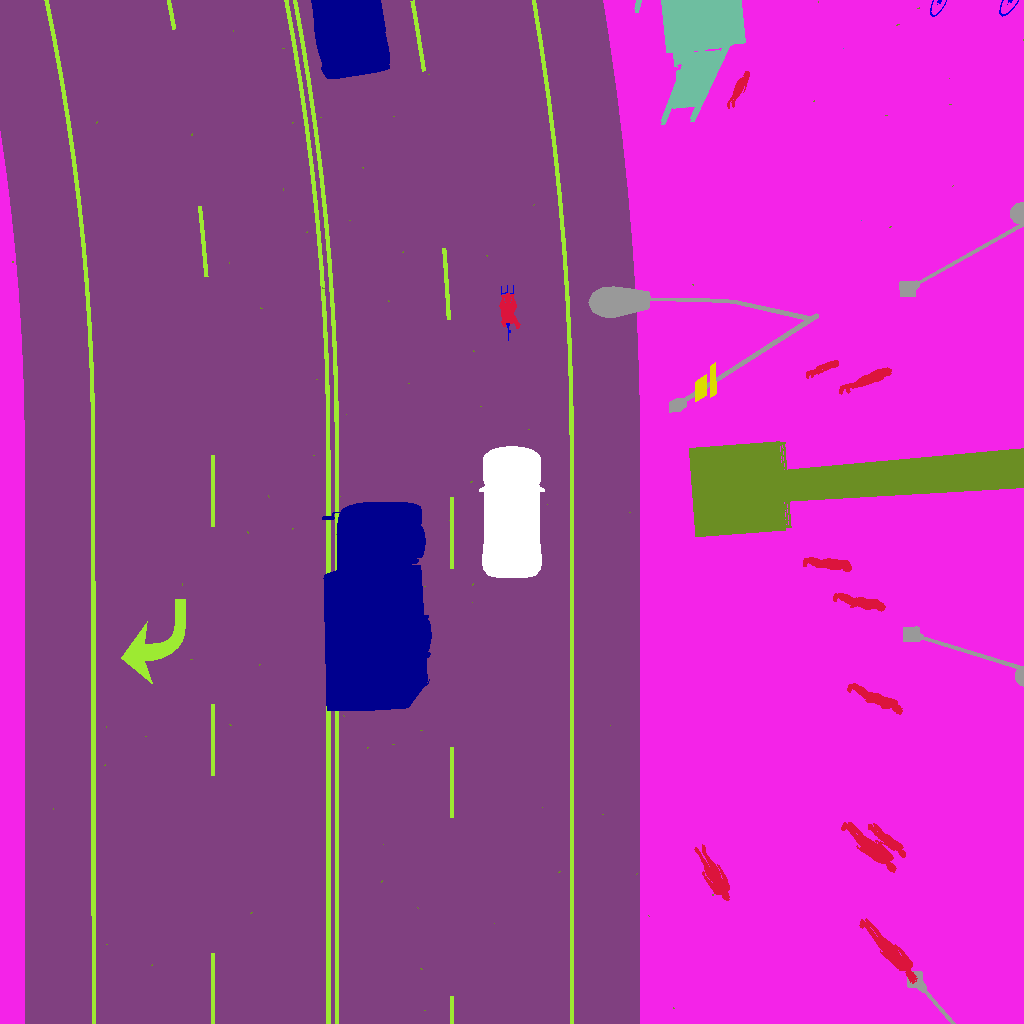}
    \vspace{-1cm}
    \caption{\textcolor{white}{(c)}}
\end{subfigure}%
\hfill
\begin{subfigure}{\sizeFish\textwidth}
    \includegraphics[width=\textwidth]{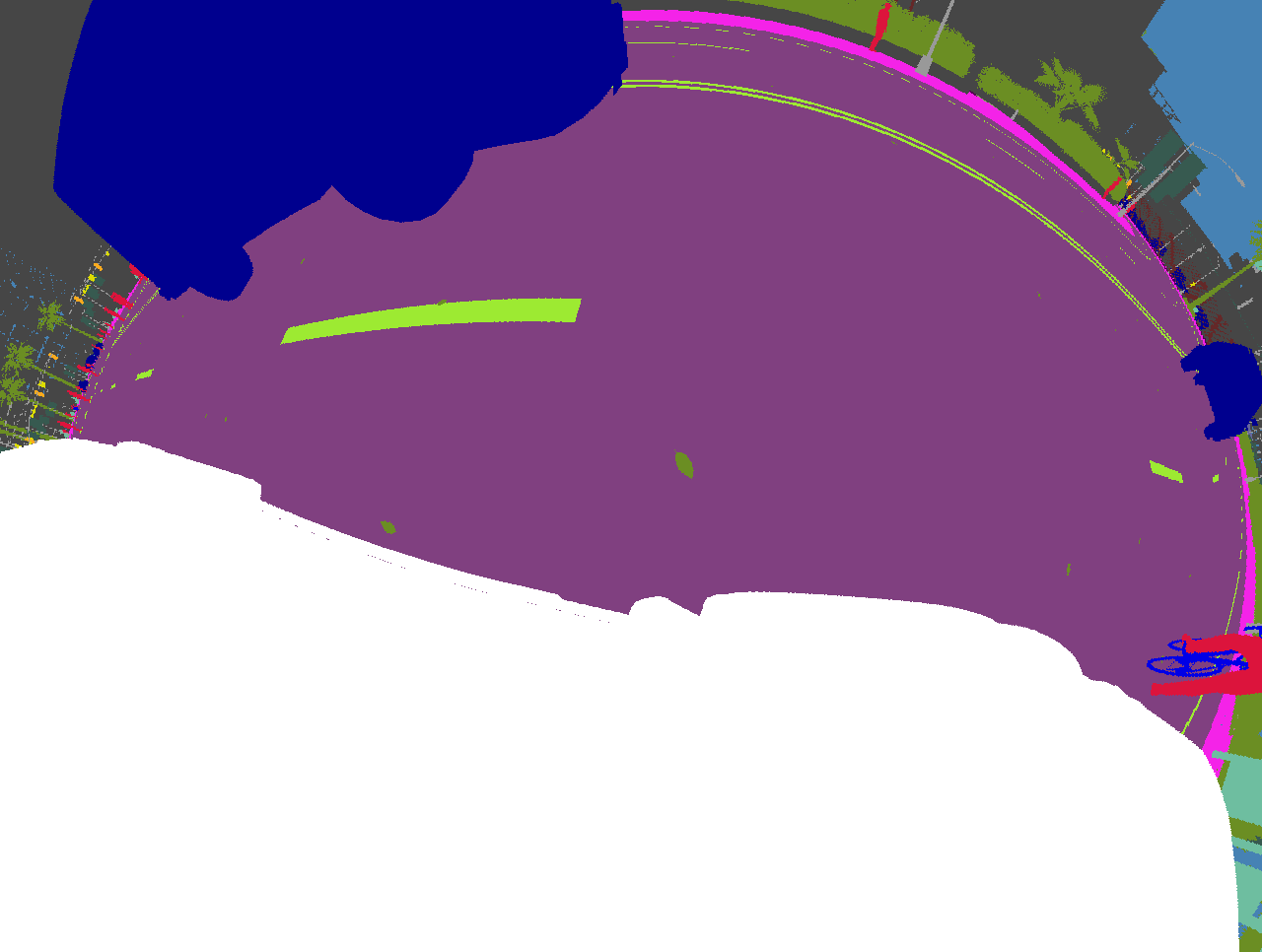}
    \vspace{-1cm}
    \caption{\textcolor{black}{(d)}}
\end{subfigure}%
\hfill
\begin{subfigure}{\sizeFish\textwidth}
    \includegraphics[width=\textwidth]{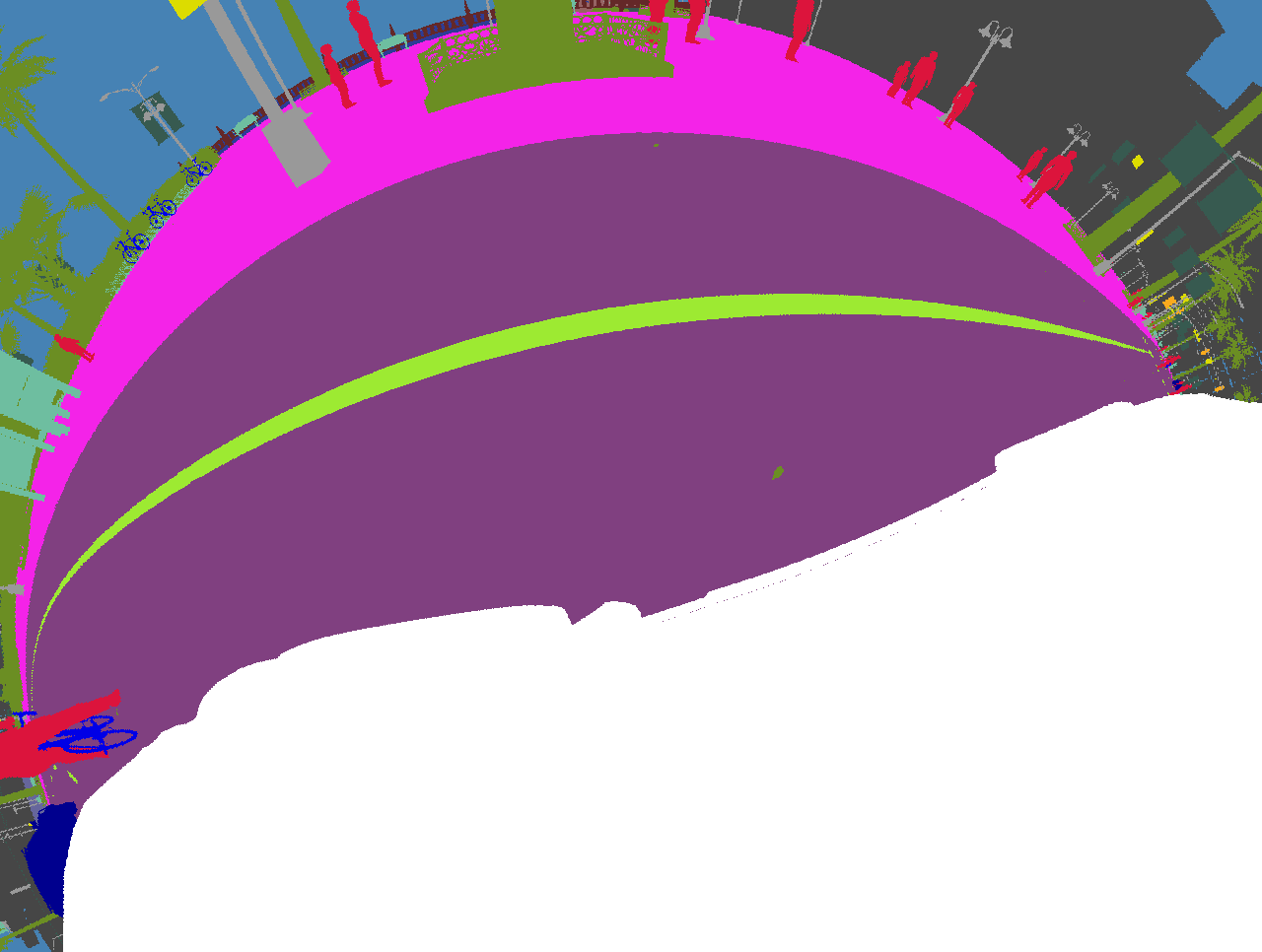}
    \vspace{-1cm}
    \caption{\textcolor{black}{(e)}}
\end{subfigure}%
\hfill
\vspace{0.07cm}


\begin{subfigure}{\sizeFish\textwidth}
    \includegraphics[width=\textwidth]{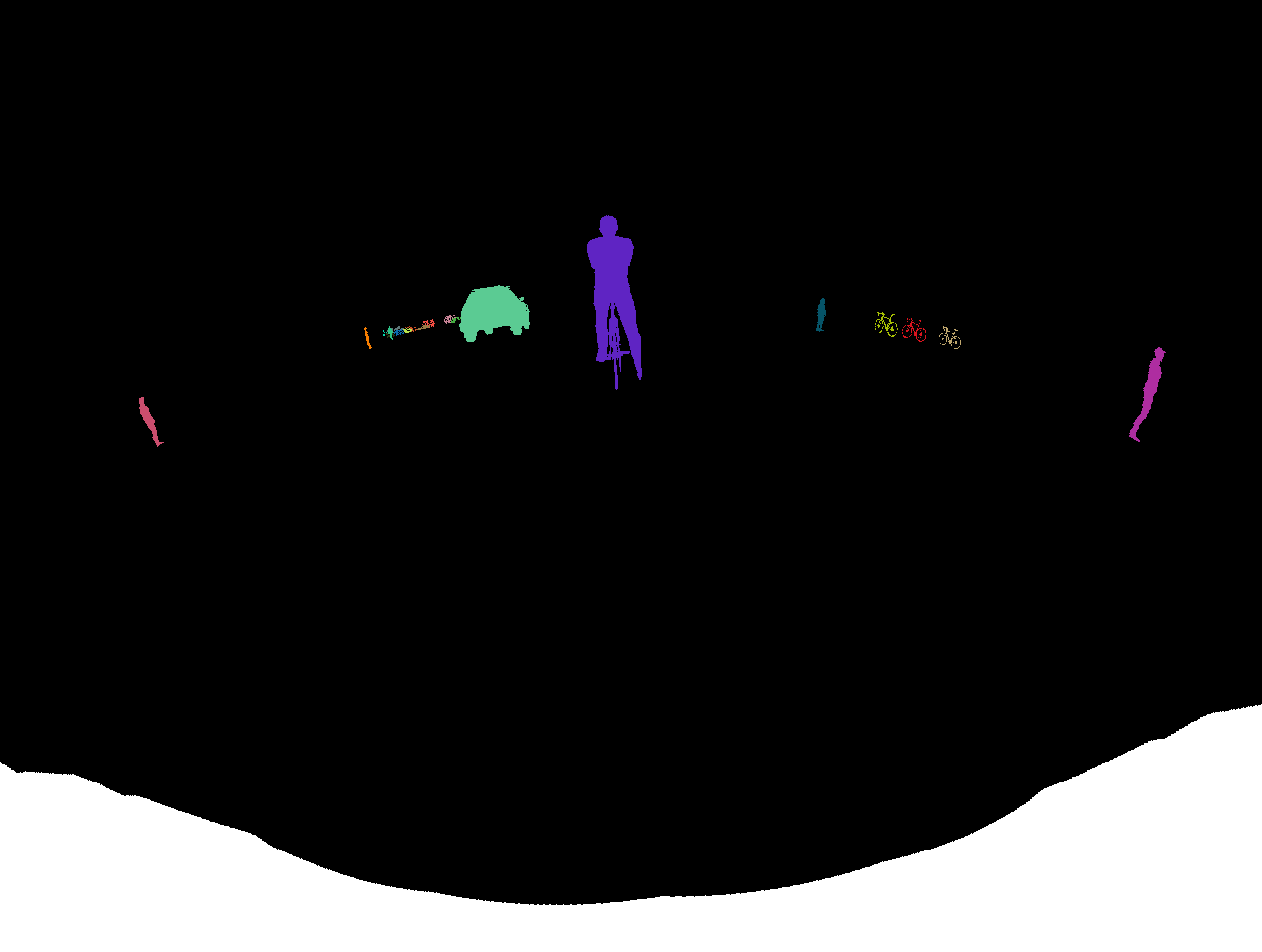}
    \vspace{-1cm}
    \caption{\textcolor{white}{(a)}}
\end{subfigure}%
\hfill
\begin{subfigure}{\sizeFish\textwidth}
    \includegraphics[width=\textwidth]{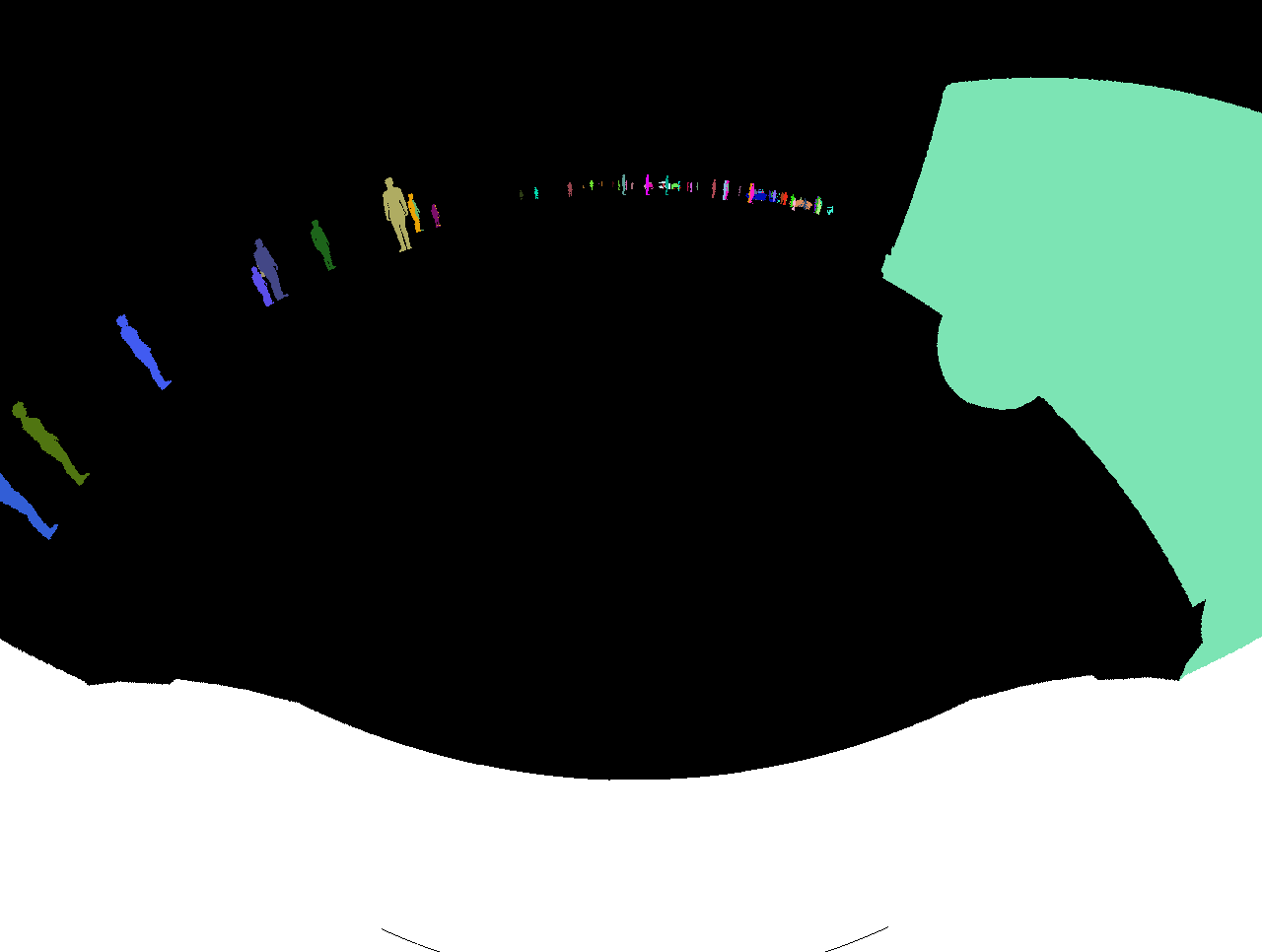}
    \vspace{-1cm}
    \caption{\textcolor{black}{(b)}}
\end{subfigure}%
\hfill
\begin{subfigure}{\sizeBEV\textwidth}
    \includegraphics[width=\textwidth]{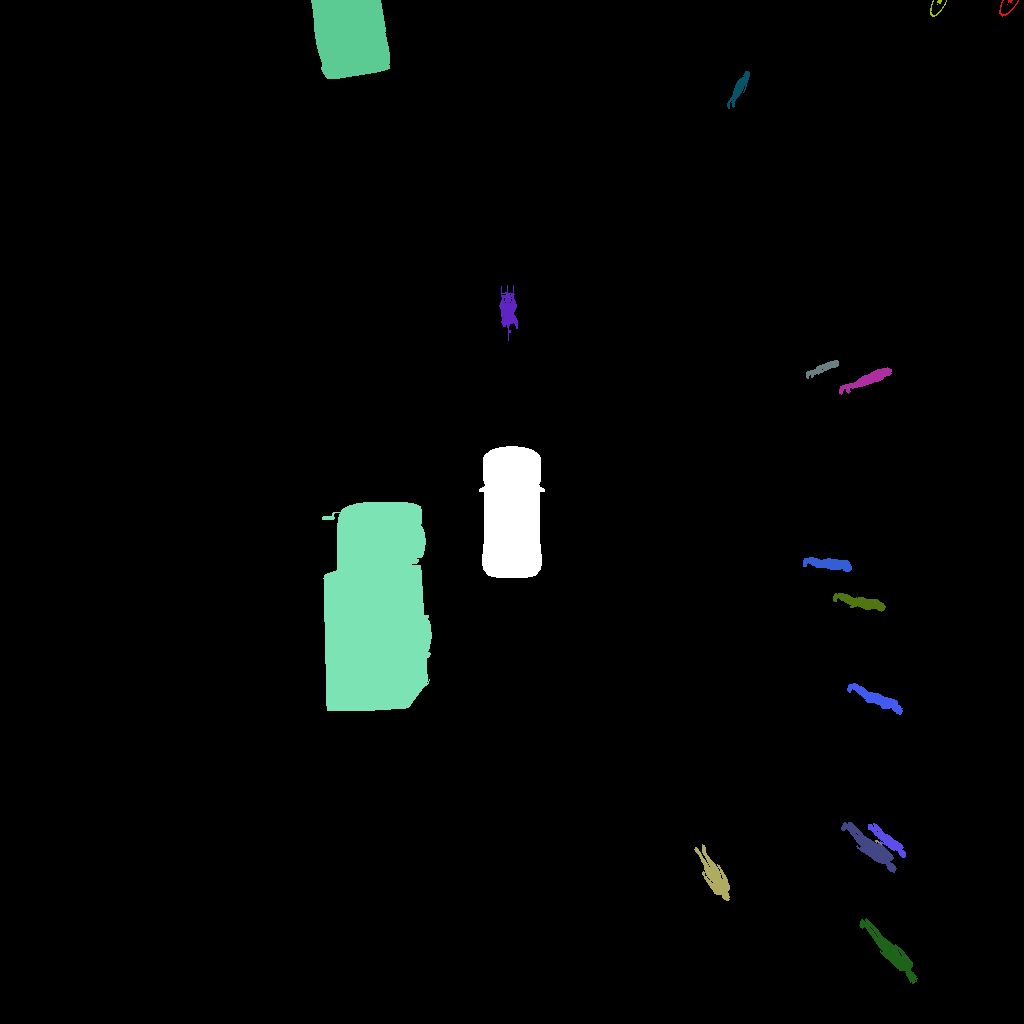}
    \vspace{-1cm}
    \caption{\textcolor{white}{(c)}}
\end{subfigure}%
\hfill
\begin{subfigure}{\sizeFish\textwidth}
    \includegraphics[width=\textwidth]{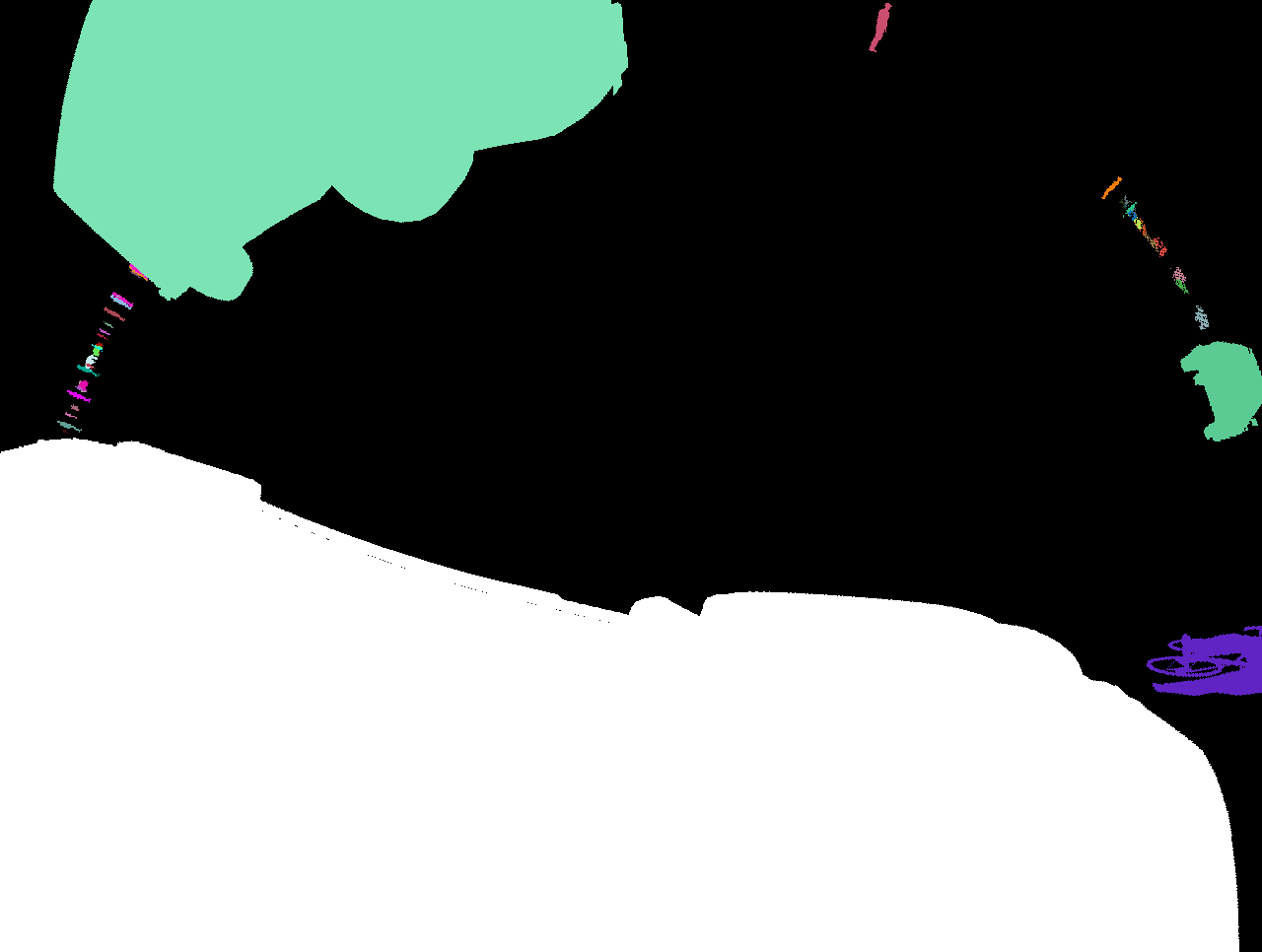}
    \vspace{-1cm}
    \caption{\textcolor{black}{(d)}}
\end{subfigure}%
\hfill
\begin{subfigure}{\sizeFish\textwidth}
    \includegraphics[width=\textwidth]{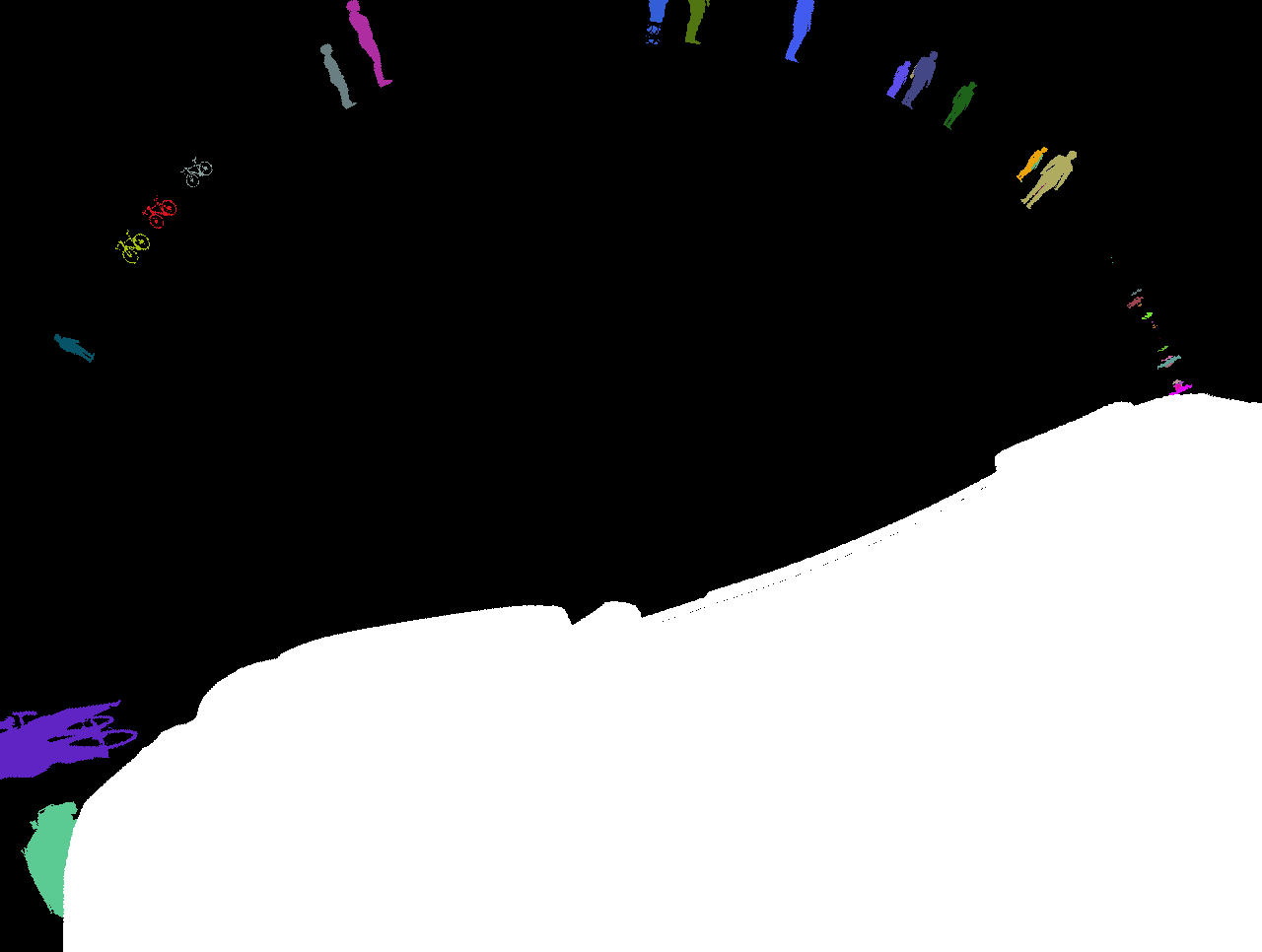}
    \vspace{-1cm}
    \caption{\textcolor{black}{(e)}}
\end{subfigure}%
\hfill
\vspace{0.07cm}


\begin{subfigure}{\sizeFish\textwidth}
    \includegraphics[width=\textwidth]{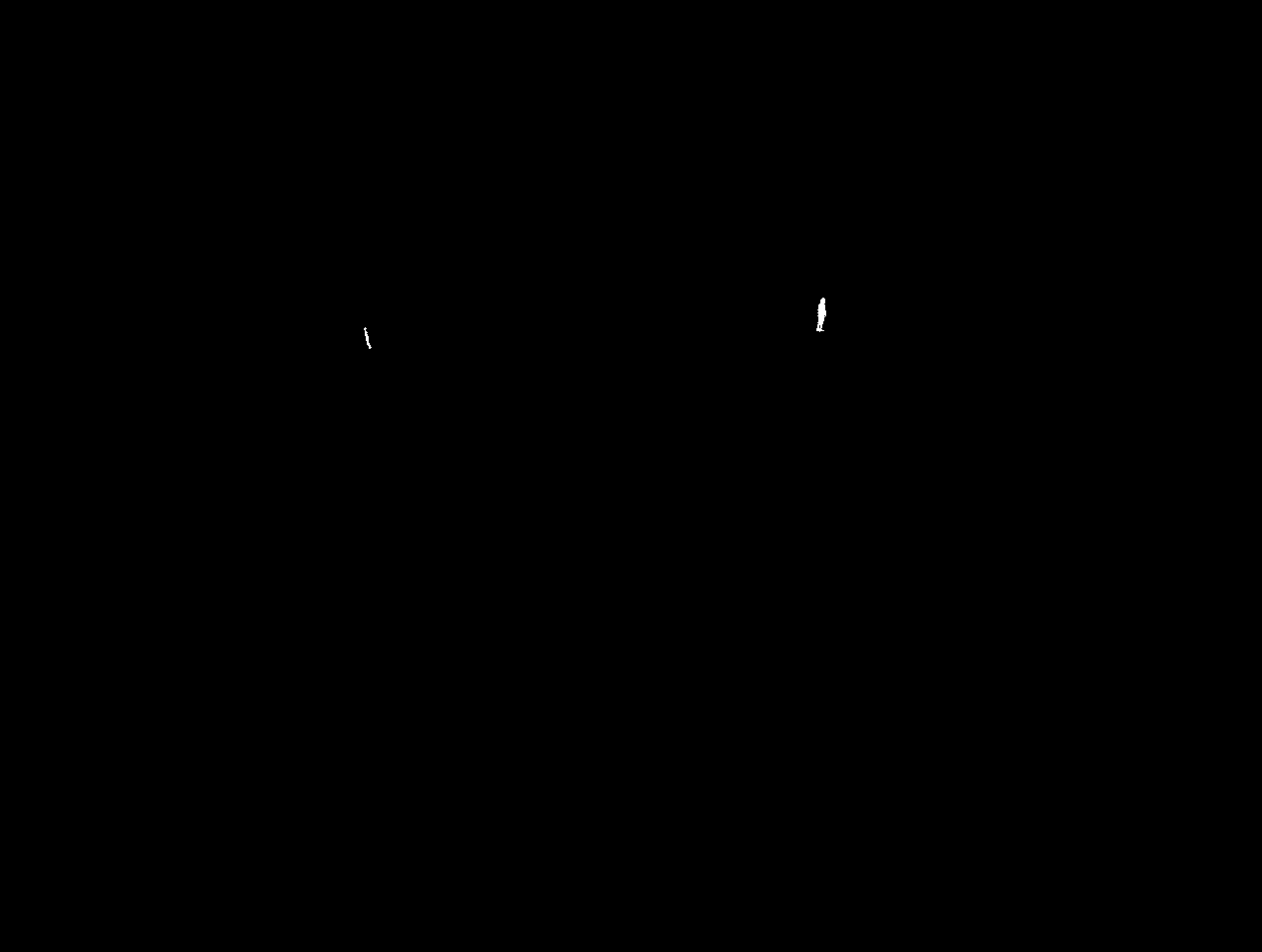}
    \vspace{-1cm}
    \caption{\textcolor{white}{(a)}}
\end{subfigure}%
\hfill
\begin{subfigure}{\sizeFish\textwidth}
    \includegraphics[width=\textwidth]{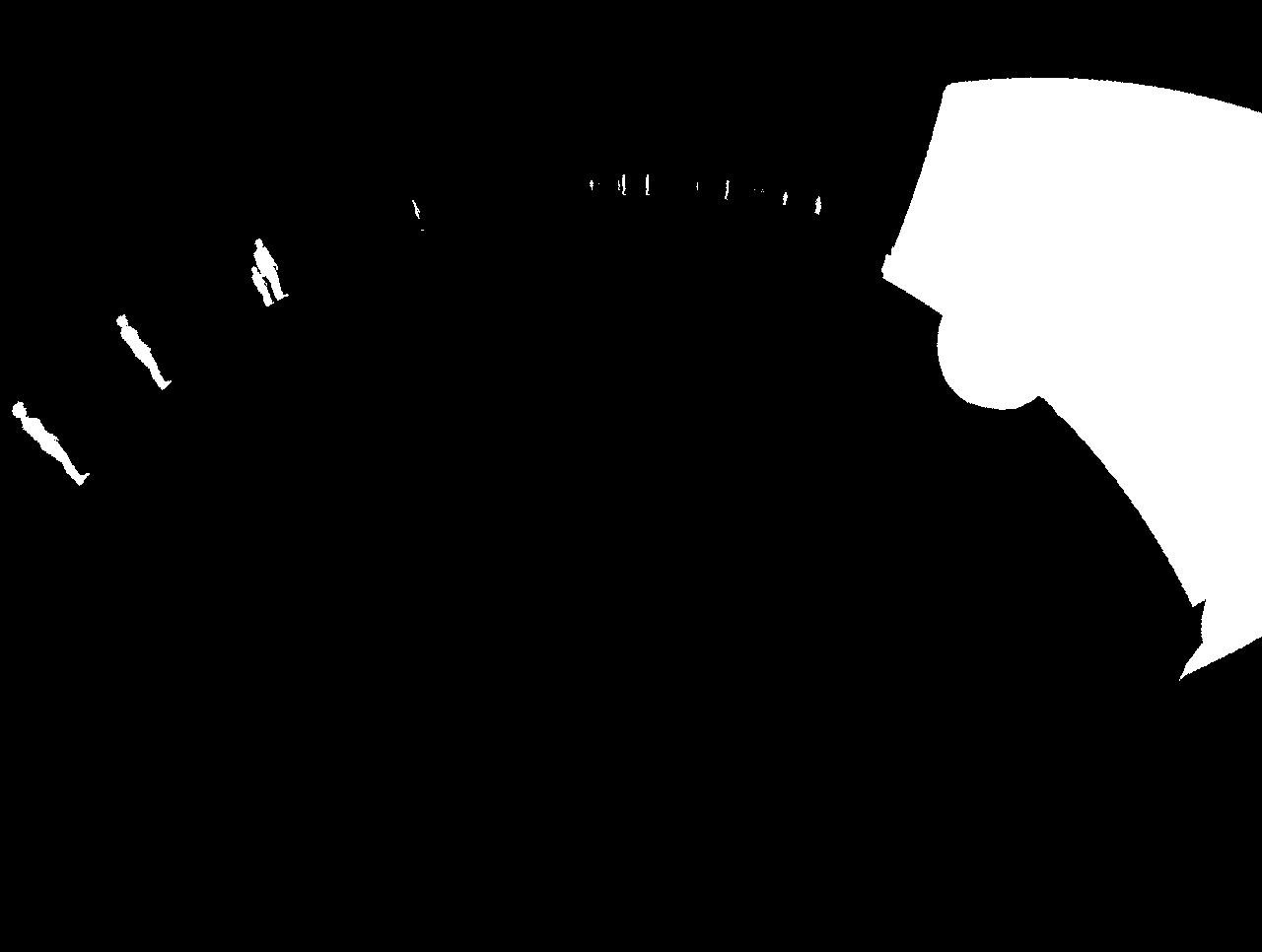}
    \vspace{-1cm}
    \caption{\textcolor{white}{(b)}}
\end{subfigure}%
\hfill
\begin{subfigure}{\sizeBEV\textwidth}
    \includegraphics[width=\textwidth]{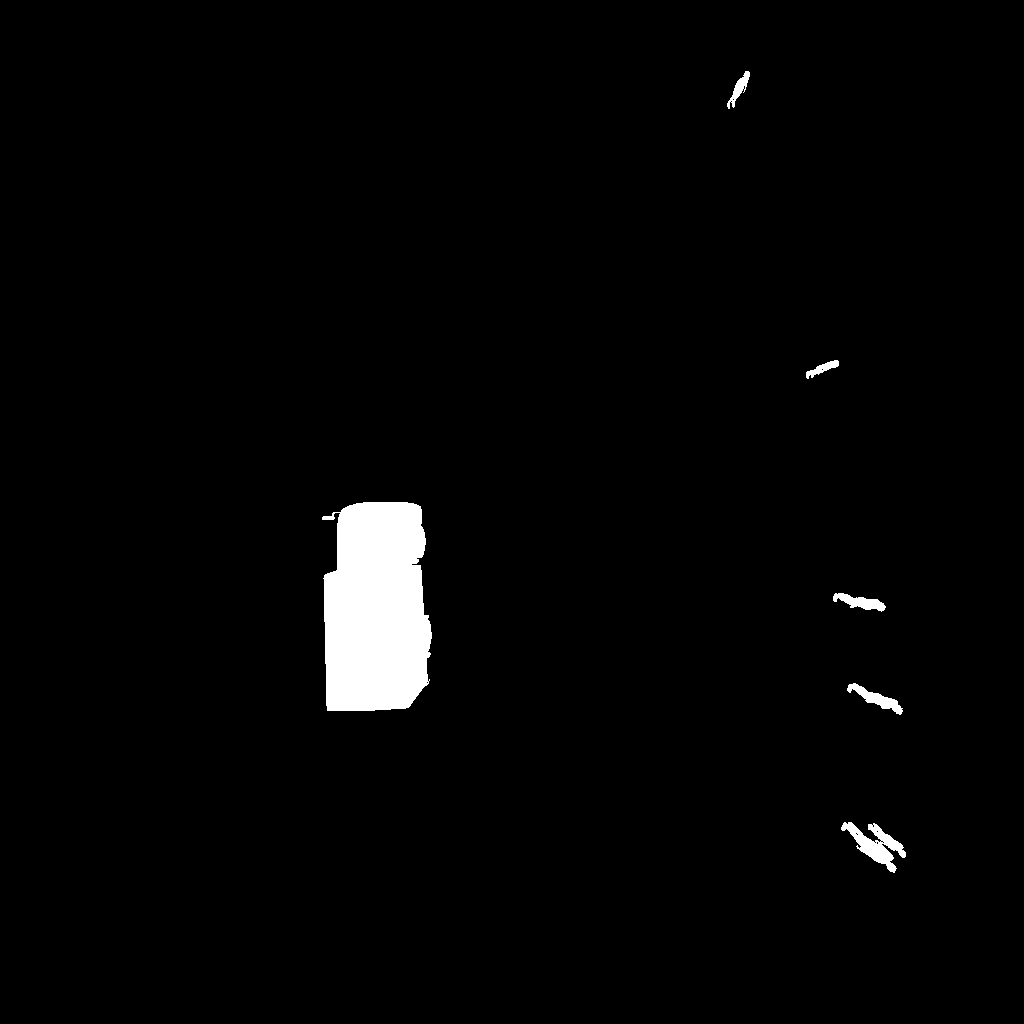}
    \vspace{-1cm}
    \caption{\textcolor{white}{(c)}}
\end{subfigure}%
\hfill
\begin{subfigure}{\sizeFish\textwidth}
    \includegraphics[width=\textwidth]{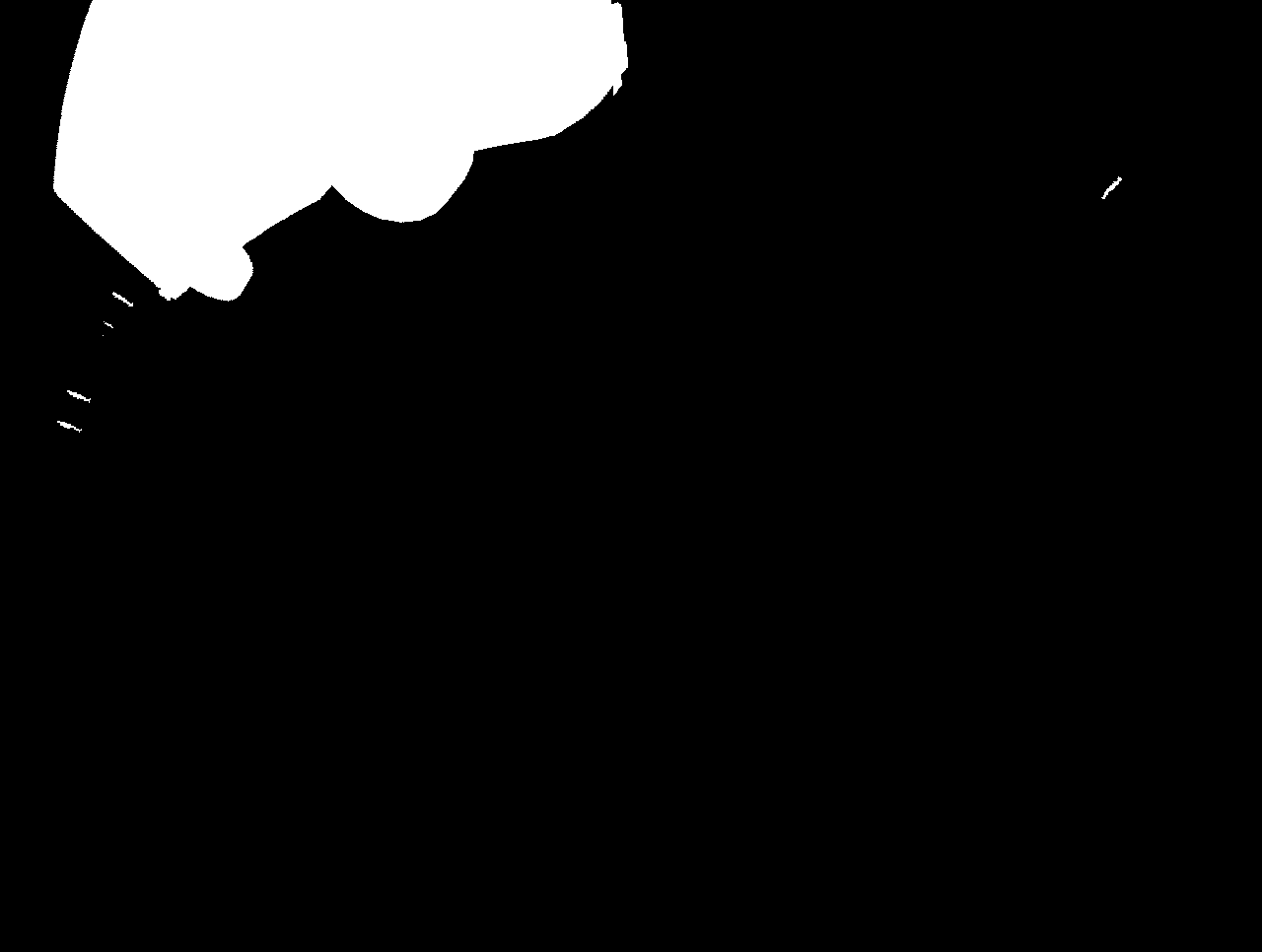}
    \vspace{-1cm}
    \caption{\textcolor{white}{(d)}}
\end{subfigure}%
\hfill
\begin{subfigure}{\sizeFish\textwidth}
    \includegraphics[width=\textwidth]{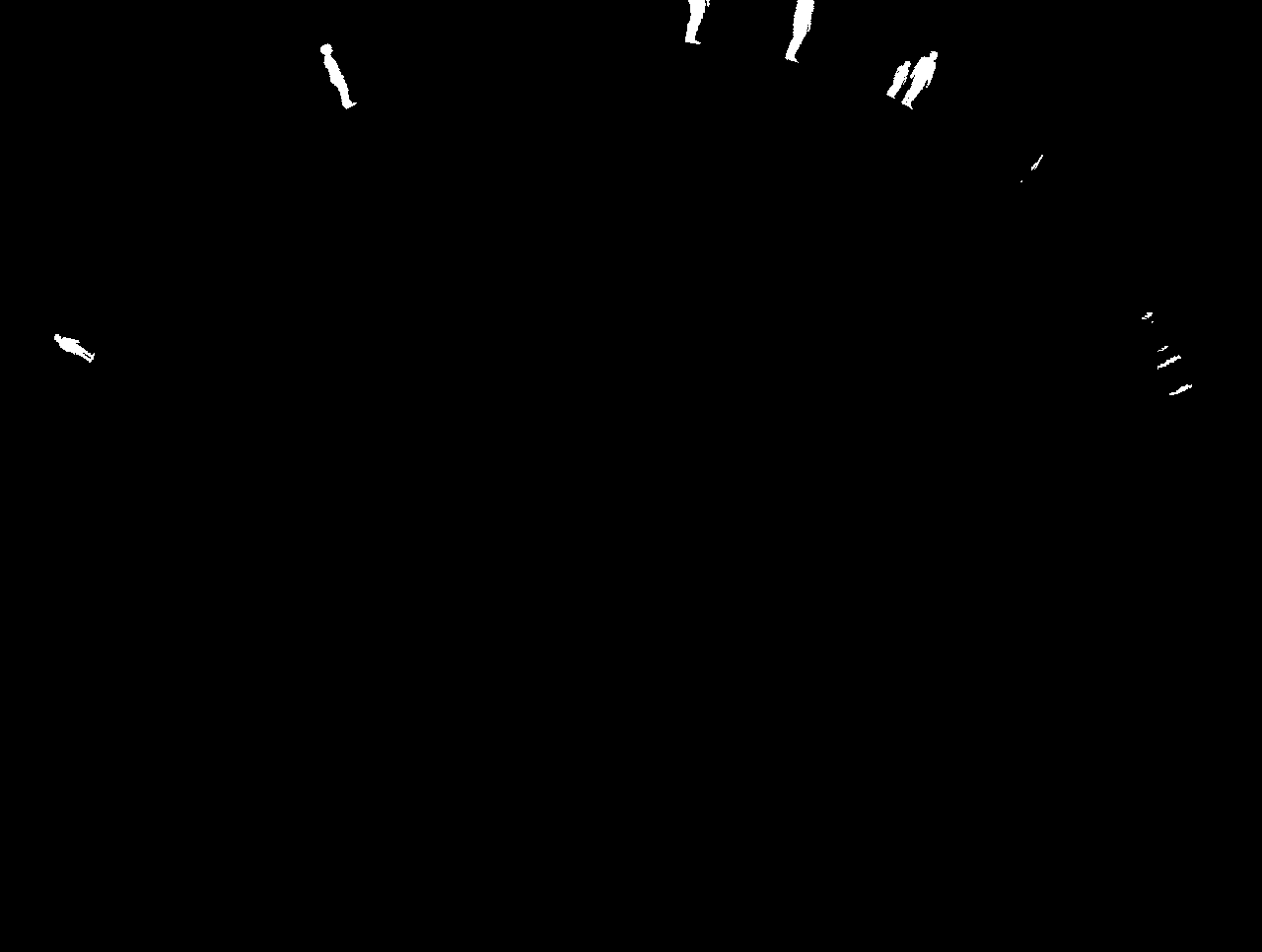}
    \vspace{-1cm}
    \caption{\textcolor{white}{(e)}}
\end{subfigure}%
\hfill
\vspace{0.07cm}


\begin{subfigure}{\sizeFish\textwidth}
    \includegraphics[width=\textwidth]{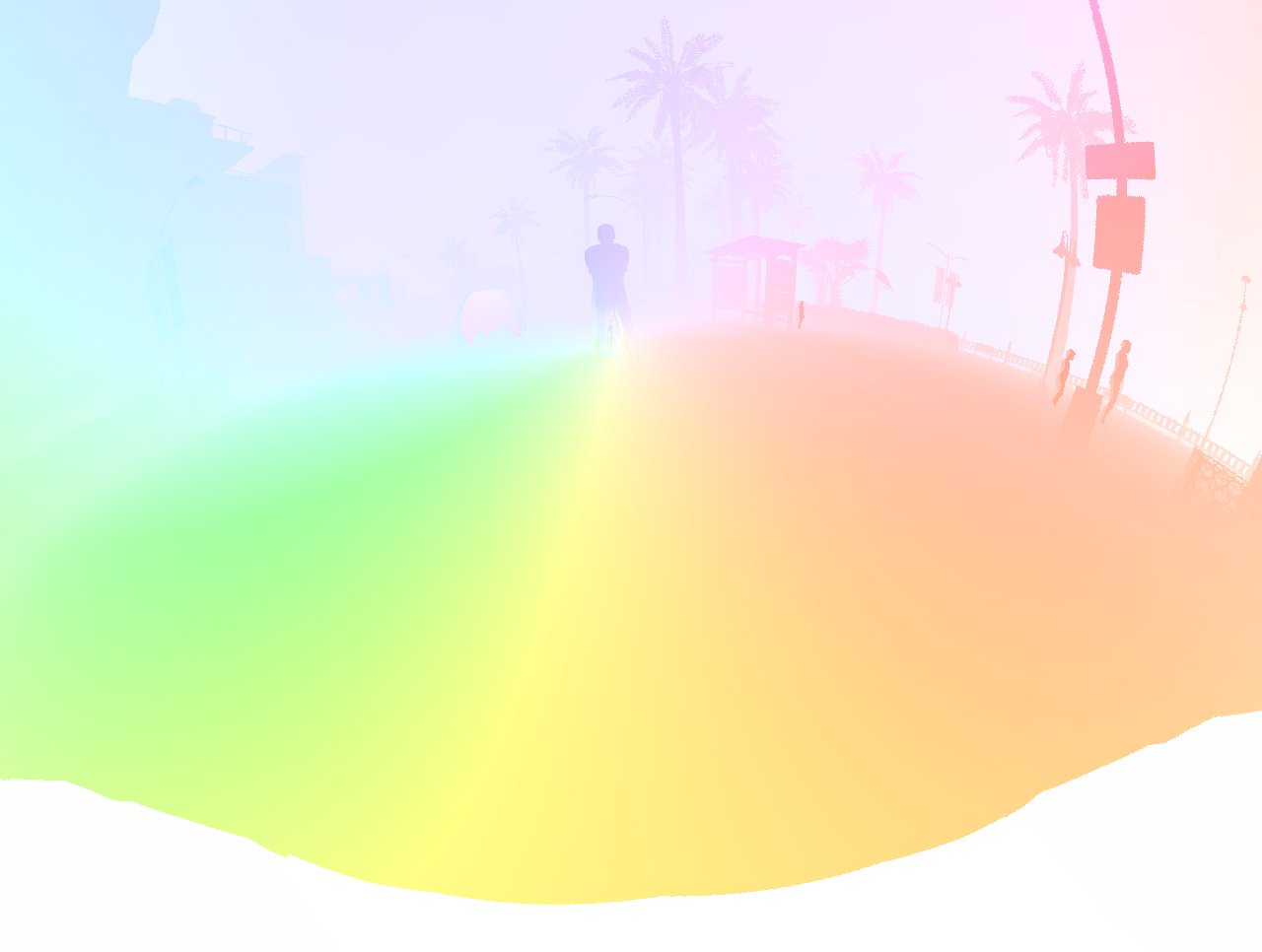}
    \vspace{-1cm}
    \caption{\textcolor{black}{(a)}}
\end{subfigure}%
\hfill
\begin{subfigure}{\sizeFish\textwidth}
    \includegraphics[width=\textwidth]{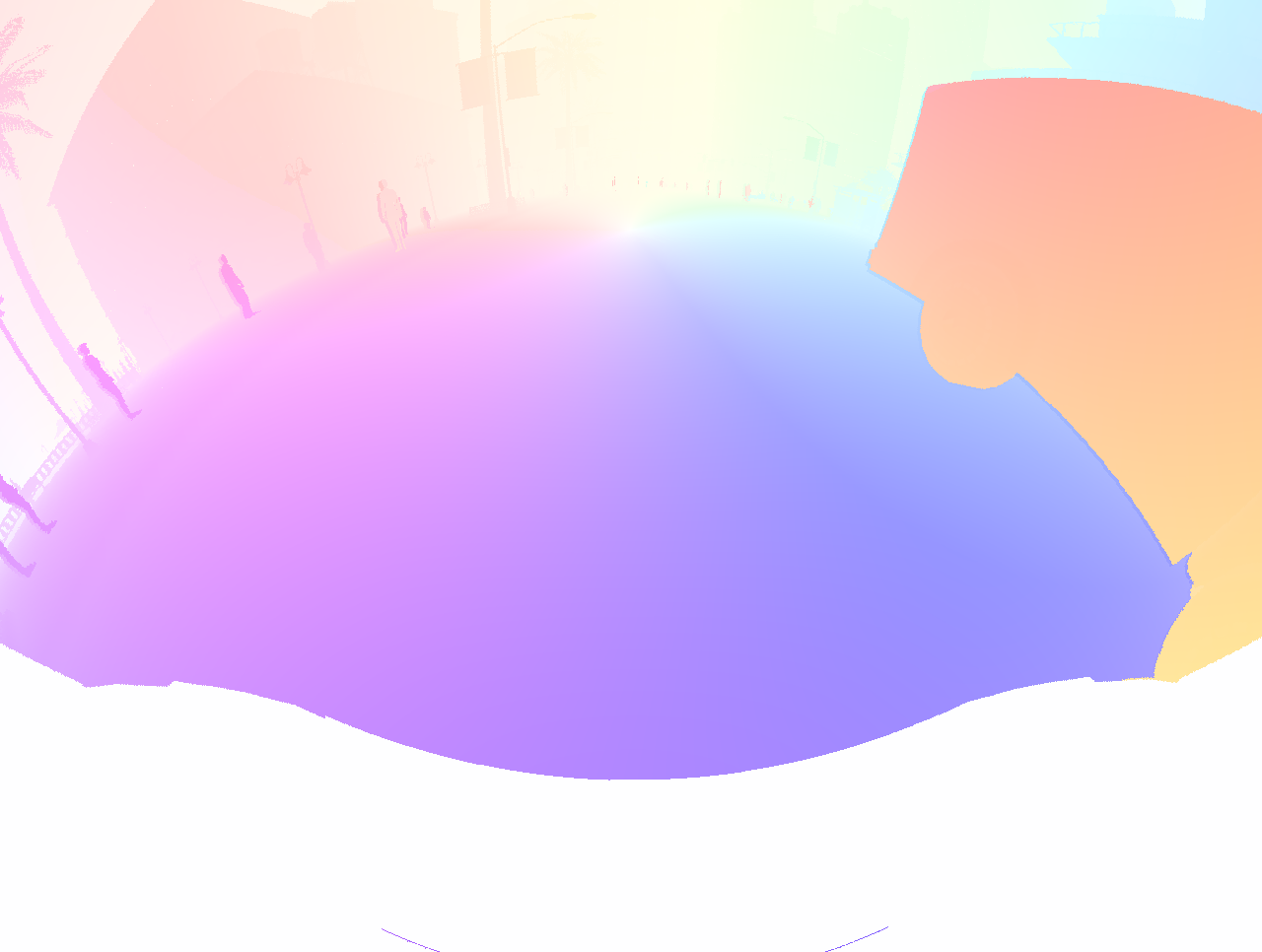}
    \vspace{-1cm}
    \caption{\textcolor{black}{(b)}}
\end{subfigure}%
\hfill
\begin{subfigure}{\sizeBEV\textwidth}
    \includegraphics[width=\textwidth]{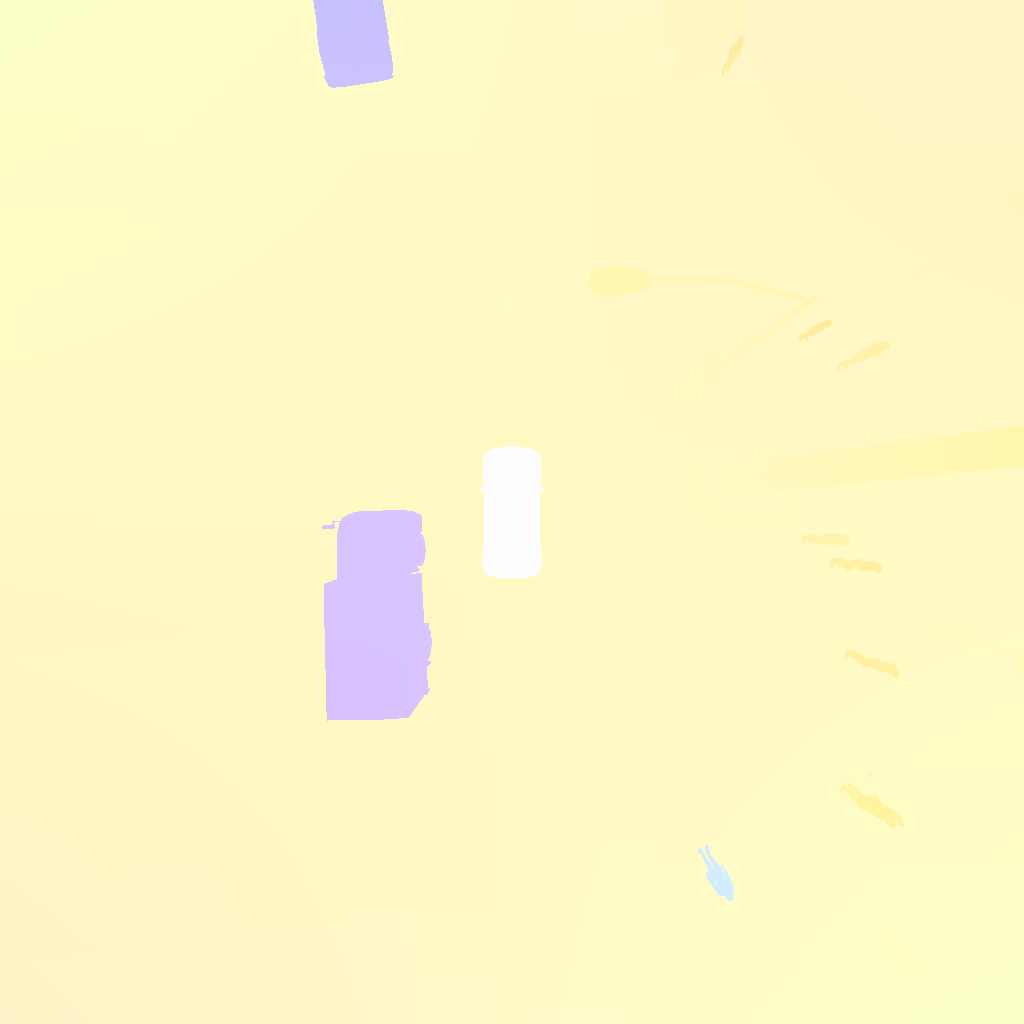}
    \vspace{-1cm}
    \caption{\textcolor{black}{(c)}}
\end{subfigure}%
\hfill
\begin{subfigure}{\sizeFish\textwidth}
    \includegraphics[width=\textwidth]{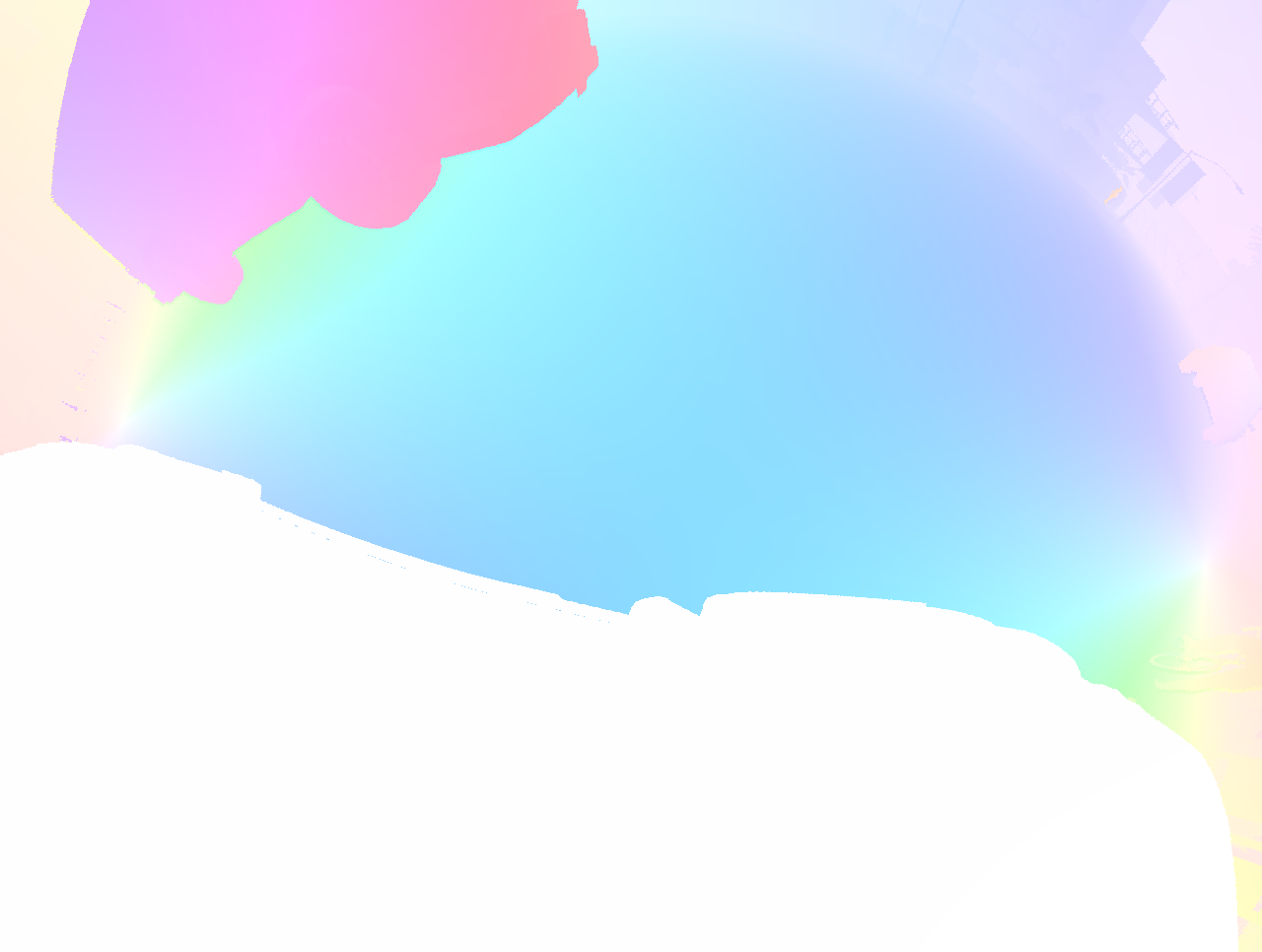}
    \vspace{-1cm}
    \caption{\textcolor{black}{(d)}}
\end{subfigure}%
\hfill
\begin{subfigure}{\sizeFish\textwidth}
    \includegraphics[width=\textwidth]{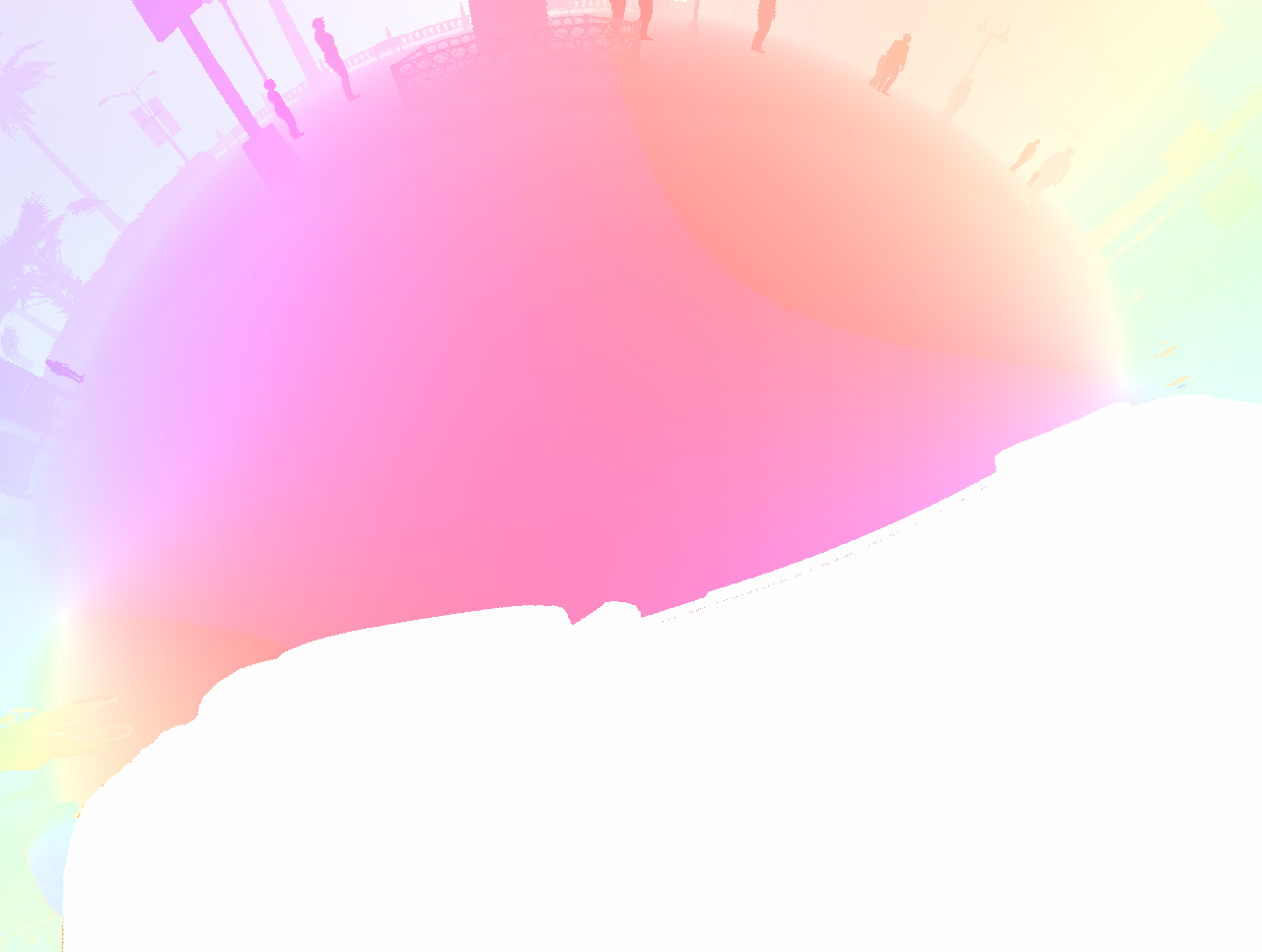}
    \vspace{-1cm}
    \caption{\textcolor{black}{(e)}}
\end{subfigure}%
\hfill
\vspace{0.07cm}


\begin{subfigure}{\sizeFish\textwidth}
    \includegraphics[width=\textwidth]{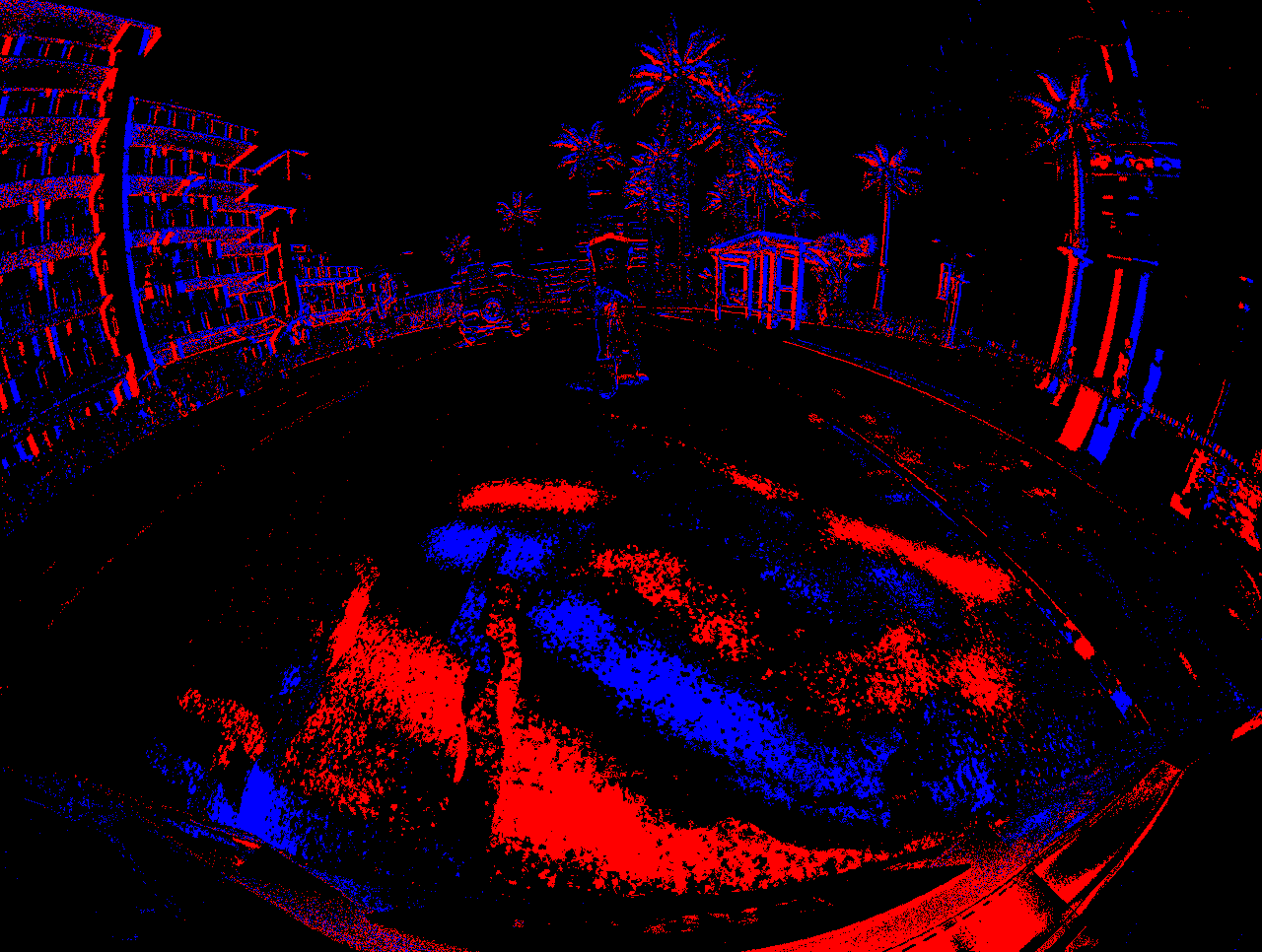}
    \vspace{-1cm}
    \caption{\textcolor{white}{(a)}}
\end{subfigure}%
\hfill
\begin{subfigure}{\sizeFish\textwidth}
    \includegraphics[width=\textwidth]{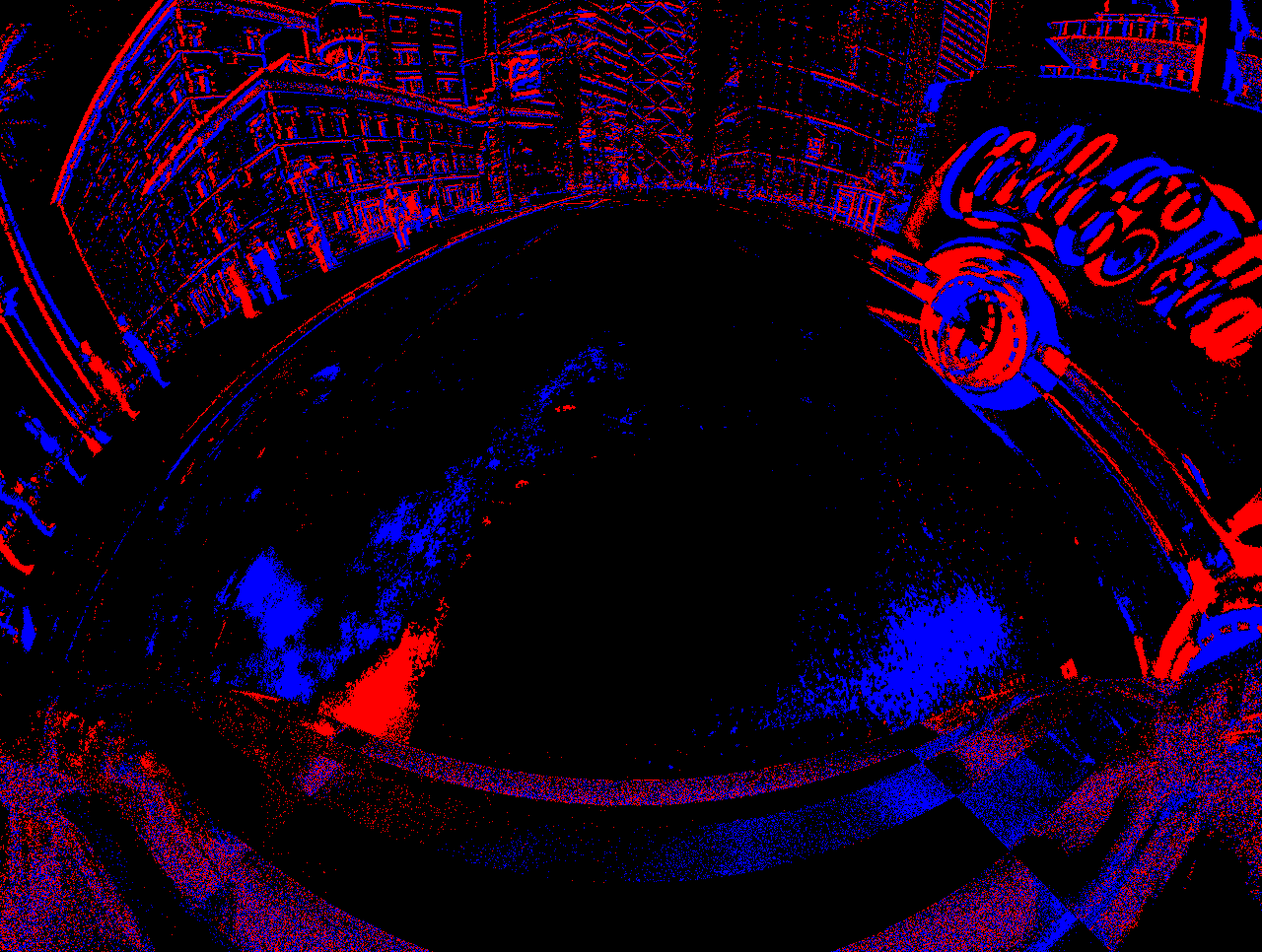}
    \vspace{-1cm}
    \caption{\textcolor{white}{(b)}}
\end{subfigure}%
\hfill
\begin{subfigure}{\sizeBEV\textwidth}
    \includegraphics[width=\textwidth]{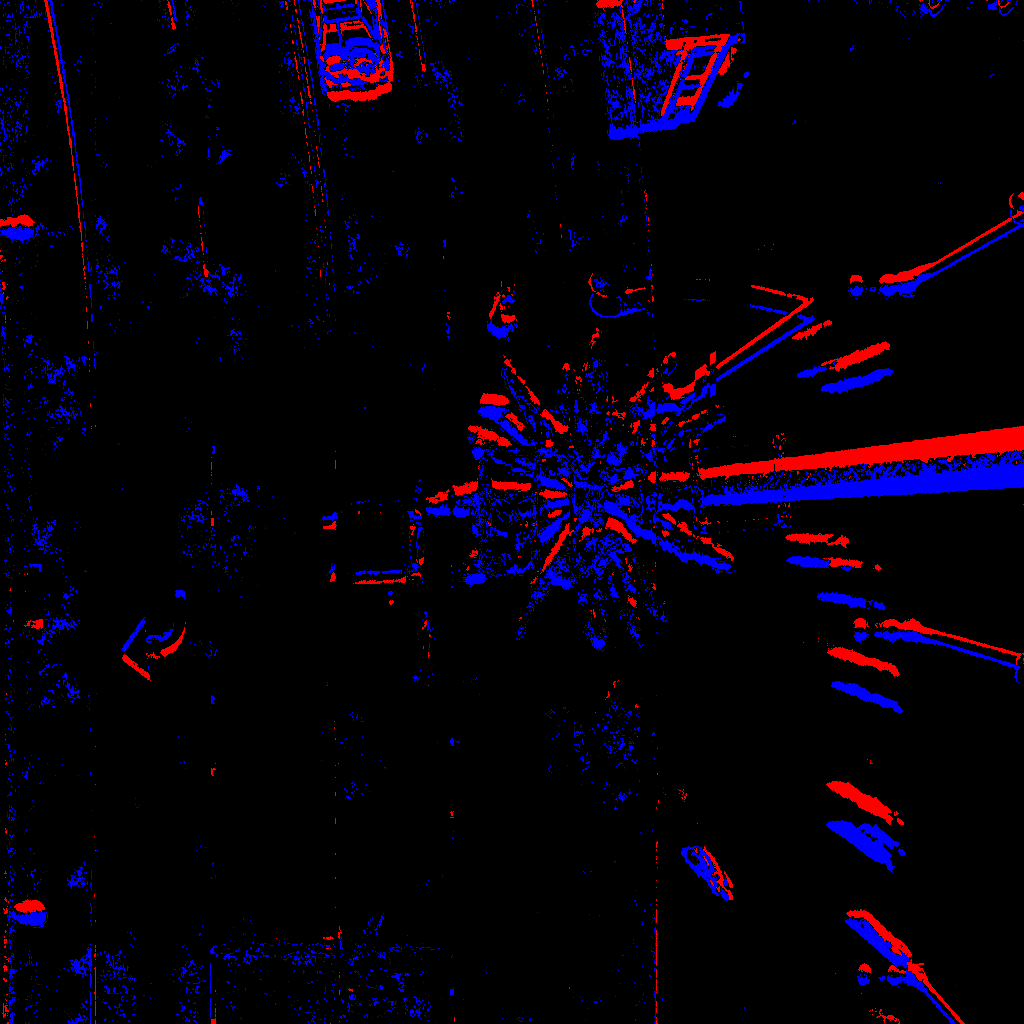}
    \vspace{-1cm}
    \caption{\textcolor{white}{(c)}}
\end{subfigure}%
\hfill
\begin{subfigure}{\sizeFish\textwidth}
    \includegraphics[width=\textwidth]{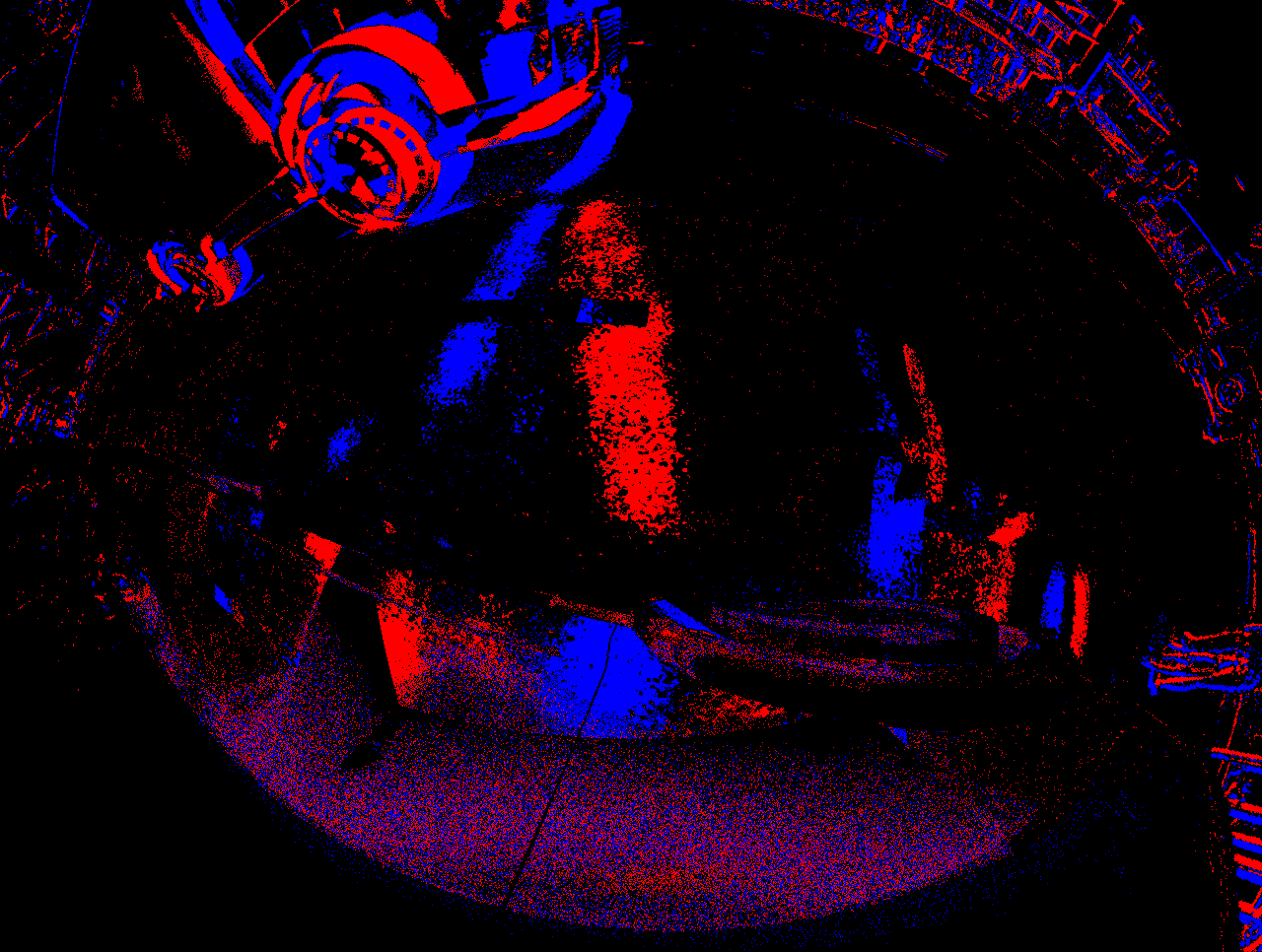}
    \vspace{-1cm}
    \caption{\textcolor{white}{(d)}}
\end{subfigure}%
\hfill
\begin{subfigure}{\sizeFish\textwidth}
    \includegraphics[width=\textwidth]{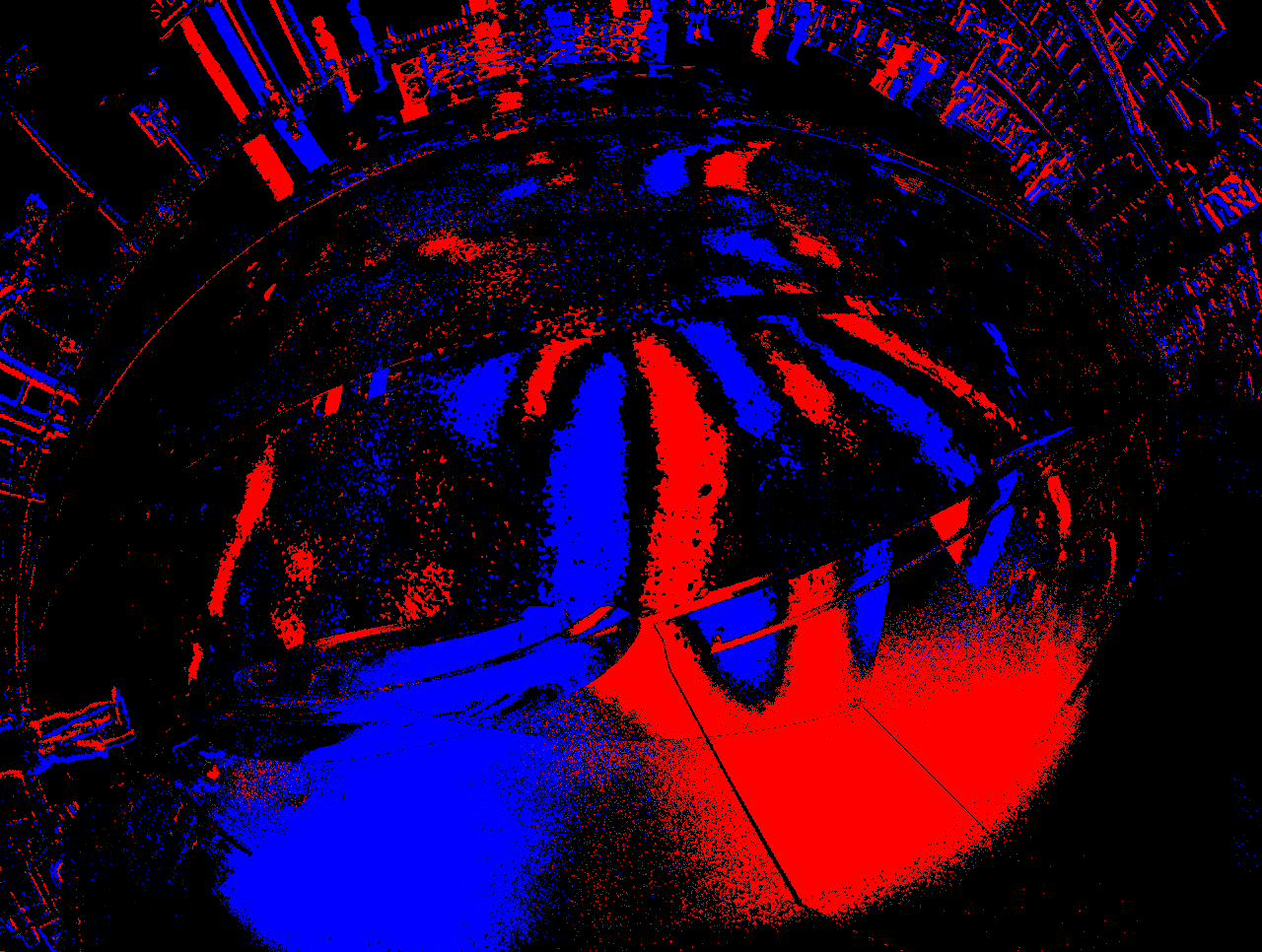}
    \vspace{-1cm}
    \caption{\textcolor{white}{(e)}}
\end{subfigure}%
\hfill

\caption{\textbf{All surround-view fisheye images and the BEV image with corresponding ground truths from a single sample.} Rows in order: RGB image pairs, Depth maps, Semantic segmentation, Instance segmentation, Motion segmentation, Optical flow (color coded), Events (positive \& negative). Cameras are marked (a) Front, (b) Rear, (c) Top view, (d) Left, (e) Right.}
    \label{One_capture}
\end{figure*}

We attribute a random color to each bounding box to obtain the instance segmentation (see \figurename~\ref{fig:teaser} and \figurename~\ref{One_capture}). The color will persist in all images captured during the current recording session.

\subsection{Motion Segmentation} \label{sec:Motion Segmentation}

Motion segmentation is the segmentation of all dynamic objects in the scene that underwent a movement in the world reference. We consider that an object has moved when the distance between the position of this object in the frames t-1 and t is superior to a threshold. Since the positions in the simulator are very precise, not using a threshold will give us a noisy motion segmentation. We will be considering in this case the small motions of the objects which will end up considering almost all objects likely to move as moving objects. To construct the ground truth of this segmentation, we used the instance segmentation and the transformation matrices of each dynamic object in the scene. We then compute the distances traveled by each object between the two frames. Since each object has a unique id and corresponds to a unique label in the instance segmentation, we can then build the motion segmentation by selecting the object in the instance segmentation that will also be included or not in the motion segmentation depending on the traveled distances.
\rd{We provide motion masks for objects that have traveled more than 0.5 meters for direct use. We also provide text files including the precise motion obtained. This allows researchers to use different threshold values depending on the use case at hand.}\par
\subsection{Optical Flow} \label{sec:Optical Flow}

Next, we explain how we compute the optical flow analytically using the data extracted from the simulator. First, we compute the scene flow, and then we project it to the image plane of all representations (perspective and fisheye) to obtain the optical flow, as described in \figurename~\ref{fig:SynWoodScape_diagram}. In this manner, we are able to provide a very precise flow information at sub-pixel level. Similar to instance segmentation, we compute the 3D point cloud of all objects in the scene separately by separating dynamic ones from static ones. Since we can extract the positions and rotations of all objects from the simulator, we can compute the transformation matrices in the 3D reference between two frames. Then, we get the scene flow by applying the transformation matrices of the movements to the point cloud of dynamic objects and the inverse of the transformation matrix of the camera movement to all 3D point clouds (dynamic and static objects). Next, we project this 3D point cloud before and after being moved into the images. This means that we have the 2D coordinates of each pixel in both frames. The vectors of movement are then constructed by each couple of these 2D coordinates representing the optical flow. These vectors can be displayed using color-coding (see \figurename~\ref{fig:teaser} and \figurename~\ref{One_capture}). We provide optical flow for all modalities using this process since we have all the calibration parameters.\par
\subsection{Event Camera} \label{sec:Event Camera}
CARLA Simulator provides event camera signals for perspective images in the form $e=(x, y, t, pol)$, where $e$ is the event triggered at pixel $(x, y)$ at timestamp $t$ with the polarity $pol$. The polarity of the event is positive when the brightness increases and negative otherwise. We compute the fisheye event camera signals using the lookup tables that allow us to map from cubemap images to fisheye images. For each event that occurred in the cubemap representation at $(x, y)$, if the pixel at $(x, y)$ is used to create the fisheye image, we compute the corresponding pixel coordinates in the fisheye image using the lookup tables. The corresponding event information $t$ and $pol$ are then assigned. Similar data structures for the perspective representation generated by the CARLA Simulator are then created for the fisheye representation and stored into NumPy array files. \figurename~\ref{fig:teaser} and \figurename~\ref{One_capture} show examples of the fisheye event signal as an RGB image where blue represents positive polarity and red is the negative one.\par
\subsection{Bird's Eye View}
\label{subsec:bev}

Behavior Prediction and Planning are generally made in the top view (or bird’s-eye-view) in a typical AD stack, \rd{due to its effective capability of representing the full scene in all directions in one representation, thus providing} most of the information an autonomous vehicle needs can be conveniently represented with the top view. 
The top view map is based on images acquired by multiple cameras looking in different directions of the vehicle at the same time. 
For the dynamic participants, we introduce the concept of instances. This makes it simple to use prior knowledge of dynamic objects to forecast behavior. Cars, for example, follow a specific motion model and have constrained patterns of future trajectory, whereas pedestrians move more randomly.
\rd{The conventional bird's eye view representation usually ignores height information. We argue that height information is very important in a lot of use cases. For instance, parking over curb scenarios requires the knowledge of the curb's height because parking on very high curbs is not possible. Speed bumps as well allow for slow driving. Therefore, unlike WoodScape, we provide height maps to enable the prediction of such objects and thus help research in that area.} \par
\begin{table}[t]
    \captionsetup{singlelinecheck=false, font=footnotesize, belowskip=0pt}
\centering
\caption{\textbf{Ablation study of OmniDet \cite{kumar2021omnidet} on WoodScape and SynWoodScape datasets}. S$^\dagger$ indicates test on the synthetic dataset SynWoodScape with {training} on the R (real-world) WoodScape dataset. R$^\dagger$ indicates test on the real-world WoodScape dataset with {training} on the S (synthetic) SynWoodScape dataset. R+S indicates mixed training of real-world and synthetic datasets.}
\label{tab:syn_vs_real}
\begin{adjustbox}{width=\columnwidth}
\begin{tabular}{@{}l|lll|lll@{}}
\toprule
\multicolumn{1}{l|}{\textit{Datasets}} & \multicolumn{3}{c|}{\textit{WoodScape}} & \multicolumn{3}{c}{\textit{SynWoodScape}} \\ 
\midrule
Train/Test & \multicolumn{1}{c}{R} & \multicolumn{1}{c}{S$^\dagger$} & R+S & \multicolumn{1}{c}{R$^\dagger$} & \multicolumn{1}{c}{S} & R+S \\
\midrule
Depth Est. (RMSE in meters)     & 1.332 & 2.401 & 1.479 & 2.393 & 1.448 & 1.396 \\
Semantic Seg. (mIoU in \%) & 76.6  & 71.7  & 76.2  & 72.1  & 78.2  & 77.8  \\
Motion Seg. (mIoU in \%)   & 75.3  & 69.5  & 74.5  & 70.7  & 76.8  & 75.1  \\
Object Det. (mAP in \%)   & 68.4  & 61.2  & 67.7  & 61.9  & 69.2  & 68.5  \\
\bottomrule
\end{tabular}
\end{adjustbox}
\end{table}

\section{Experiments}

\subsection{Real vs. Synthetic Baseline performance}

In \tablename~\ref{tab:syn_vs_real}, we establish a baseline benchmark for the SynWoodScape dataset as an ablation study using the OmniDet \rd{framework \cite{kumar2021omnidet}, which is a surround-view cameras based multi-task visual perception network for AD evaluated on the WoodScape and SynWoodScape datasets. An important aspect of this particular ablation study entails evaluating the need for domain transfer, establishing a baseline for the community, and evaluating our framework to test the model generalization capabilities. Because of the differences in synthetic and real-world data, listed perception tasks do not yield quantitatively desired results when applied directly to real-world data, necessitating the domain adaptation phase. 
Initially, we train on the WoodScape and test it on the SynWoodScape to establish a baseline for the domain transfer. Later, we mix both datasets and train on them jointly to set up a quantitative baseline for these datasets. Finally, we train on SynWoodScape which serves as a standalone baseline, and also evaluate it on WoodScape to measure the deviation of the domain gap.}
\rd{We perform such an ablation study on 4 tasks as reported in \tablename~\ref{tab:syn_vs_real} where the 4-task model is trained jointly.} 

\rd{RMSE has been used as an accuracy metric for depth estimation, while mIoU is used for semantic and motion segmentation and mAP is object detection. These metrics are standard for such tasks across the literature. We use the same data split that was done in OmniDet to be able to compare our results to OmniDet's official benchmark. The first column of results ``\textbf{R}" reports the accuracy of real data after training on real data, which corresponds to the results reported in OmniDet. When evaluated on the synthetic dataset from SynWoodScape, we obtained degraded performance as reported in  
``\textbf{S$^\dagger$}". This is expected because of the different nature of the datasets. When we used mixed training on both datasets and evaluated the real data in ``\textbf{R+S (WoodScape)}", we obtained improved performance over ``\textbf{S$^\dagger$}"; However, the accuracy is still less than ``\textbf{R}". This result demonstrates the importance of domain adaptation to be able to use jointly real and synthetic data. It is well known that deep learning is data-oriented. Therefore, annotating large datasets usually provides better performance, and this is a time and effort expensive operation. To evaluate the usage of synthetic data only with minimal effort in manual annotation, we trained the network on synthetic data only and evaluated on real scenarios as demonstrated in ``\textbf{R$^\dagger$}". To our surprise, we obtained good accuracy with an acceptable performance given the cost of annotation. However, the result is less than ``\textbf{R}", which is expected. Evaluation on synthetic data only is illustrated in ``\textbf{S}" showing the maximum performance due to the same nature of training and testing data. Finally ``\textbf{R+S (SynWoodScape)}" demonstrates that the benchmark model is not capable of making use of the new data and therefore motivates the need for domain adaptation.}

\begin{table}[t]
    \captionsetup{singlelinecheck=false, font=footnotesize, belowskip=-2pt}
\centering
\caption{\bf {Quantitative comparison of segmentation task on Top View model vs. Transformed Model.}}
\label{tab:top_view}
\begin{tabular}{@{}ll@{}}
\toprule
\textit{Model}         & \textit{Accuracy (mIoU)} \\ 
\midrule
Image Semantic Segmentation + IPM      &    61.2 \\
Top View Semantic Segmentation &    76.5 \\ 
\bottomrule
\end{tabular}
\end{table}

\subsection{Top View Segmentation}
\label{subsec:top-view}

We ablate the OmniDet \cite{kumar2021omnidet} on the top view dataset and establish a baseline performance in \tablename~\ref{tab:top_view}. Initially, we train the model for the trivial semantic segmentation task and transform it using inverse perspective mapping (IPM) for the behavior and planning stage as explained in Section \ref{subsec:bev}. \rd{This method is considered a cost-free one as it does not need annotation to be performed. However, the transformation provides an erroneous projection and object distortion. To provide better results, we attempt to train our model directly on top view projection; However, this requires annotation. In our proposed dataset, we provide such top view annotations and they have the advantage of being cost-free as well because the data is obtained from a simulator. We train our model using our synthetic top view annotations and we obtain the results shown in the table, which show significant improvement for all segmentation tasks.
We release motion masks and instance segmentation datasets to identify the dynamic objects and localize particular vehicles/instances in the top view as many of the existing approaches tend to connect multiple cars into one contiguous region.}\par
\section{Conclusion}

In this paper, we provide a synthetic dataset using surround-view fisheye cameras dedicated to AD with ground truth annotations for 10+ tasks. In addition to providing synchronized fisheye data, we provide bird's eye view data with annotations. We demonstrated the relevance of the generated synthetic data by performing baseline experiments for depth estimation, semantic segmentation, motion segmentation, and object detection as well as experiments on the same tasks using the top view. \rd{Our experiments show the benefit of the proposed dataset in terms of performance vs. cost, where cost-free synthetic data can be used for the perception of real scenarios. The results also demonstrate the need for a domain adaptation approach to fully make use of our proposed dataset, which can be done in future work.} Because our dataset is using the same configuration and calibration parameters used in the WoodScape dataset, the couple SynWoodScape/WoodScape is of great interest in the development of models dedicated to fisheye images as well as transfer learning between real and synthetic data or image-to-image translation algorithms. 



\bibliographystyle{IEEEtran}
\bibliography{IEEEfull.bib}

\begin{thebibliography}{10}
\providecommand{\url}[1]{#1}
\csname url@samestyle\endcsname
\providecommand{\newblock}{\relax}
\providecommand{\bibinfo}[2]{#2}
\providecommand{\BIBentrySTDinterwordspacing}{\spaceskip=0pt\relax}
\providecommand{\BIBentryALTinterwordstretchfactor}{4}
\providecommand{\BIBentryALTinterwordspacing}{\spaceskip=\fontdimen2\font plus
\BIBentryALTinterwordstretchfactor\fontdimen3\font minus
  \fontdimen4\font\relax}
\providecommand{\BIBforeignlanguage}[2]{{%
\expandafter\ifx\csname l@#1\endcsname\relax
\typeout{** WARNING: IEEEtran.bst: No hyphenation pattern has been}%
\typeout{** loaded for the language `#1'. Using the pattern for}%
\typeout{** the default language instead.}%
\else
\language=\csname l@#1\endcsname
\fi
#2}}
\providecommand{\BIBdecl}{\relax}
\BIBdecl

\bibitem{popperli2019capsule}
M.~P{\"o}pperli, R.~Gulagundi, S.~Yogamani, and S.~Milz, ``Capsule neural
  network based height classification using low-cost automotive ultrasonic
  sensors,'' in \emph{2019 IEEE Intelligent Vehicles Symposium (IV)}.\hskip 1em
  plus 0.5em minus 0.4em\relax IEEE, 2019, pp. 661--666.

\bibitem{eising2021near}
C.~Eising, J.~Horgan, and S.~Yogamani, ``Near-field perception for low-speed
  vehicle automation using surround-view fisheye cameras,'' \emph{IEEE
  Transactions on Intelligent Transportation Systems}, 2021.

\bibitem{kumar2020unrectdepthnet}
R.~K. Varun, S.~Yogamani, M.~Bach, C.~Witt, S.~Milz, and P.~M{\"{a}}der,
  ``{UnRectDepthNet: Self-Supervised Monocular Depth Estimation using a Generic
  Framework for Handling Common Camera Distortion Models},'' in
  \emph{{IEEE/RSJ} International Conference on Intelligent Robots and Systems,
  {IROS}}, 2020.

\bibitem{Sekkat2022comparativestudy}
A.~R. Sekkat, Y.~Dupuis, P.~Honeine, and P.~Vasseur, ``A comparative study of
  semantic segmentation of omnidirectional images from a motorcycle
  perspective,'' \emph{Scientific Reports}, vol.~12, no.~1, p. 4968, Mar 2022.

\bibitem{9800124}
A.~Dahal, V.~R. Kumar, S.~Yogamani, and C.~Eising, ``An online learning system
  for wireless charging alignment using surround-view fisheye cameras,''
  \emph{IEEE Transactions on Intelligent Transportation Systems}, pp. 1--10,
  2022.

\bibitem{rashedfisheyeyolo}
R.~Hazem, E.~Mohamed, V.~R.~K. Sistu, Ganesh~and, C.~Eising, A.~El-Sallab, and
  S.~Yogamani, ``{FisheyeYOLO: Object Detection on Fisheye Cameras for
  Autonomous Driving},'' \emph{Machine Learning for Autonomous Driving NeurIPS
  2020 Virtual Workshop}, 2020.

\bibitem{uricar2021let}
M.~Uricar, G.~Sistu, H.~Rashed, A.~Vobecky, V.~Ravi~Kumar, P.~Krizek,
  F.~Burger, and S.~Yogamani, ``Let's get dirty: Gan based data augmentation
  for camera lens soiling detection in autonomous driving,'' in
  \emph{Proceedings of the IEEE/CVF Winter Conference on Applications of
  Computer Vision}, 2021, pp. 766--775.

\bibitem{das2020tiledsoilingnet}
A.~Das, P.~K{\v{r}}{\'\i}{\v{z}}ek, G.~Sistu, F.~B{\"u}rger, S.~Madasamy,
  M.~U{\v{r}}i{\v{c}}{\'a}{\v{r}}, V.~Ravi~Kumar, and S.~Yogamani,
  ``{TiledSoilingNet: Tile-level Soiling Detection on Automotive Surround-view
  Cameras Using Coverage Metric},'' in \emph{2020 IEEE 23rd International
  Conference on Intelligent Transportation Systems (ITSC)}.\hskip 1em plus
  0.5em minus 0.4em\relax IEEE, 2020, pp. 1--6.

\bibitem{sobh2021adversarial}
I.~Sobh, A.~Hamed, V.~Ravi~Kumar, and S.~Yogamani, ``Adversarial attacks on
  multi-task visual perception for autonomous driving,'' \emph{Journal of
  Imaging Science and Technology}, vol.~65, no.~6, pp. 60\,408--1, 2021.

\bibitem{dahal2021roadedgenet}
A.~Dahal, E.~Golab, R.~Garlapati, V.~Ravi~Kumar, and S.~Yogamani,
  ``{RoadEdgeNet: Road Edge Detection System Using Surround View Camera
  Images},'' in \emph{Electronic Imaging}, 2021.

\bibitem{dhananjaya2021weather}
M.~M. Dhananjaya, V.~R. Kumar, and S.~Yogamani, ``Weather and light level
  classification for autonomous driving: Dataset, baseline and active
  learning,'' in \emph{2021 IEEE International Intelligent Transportation
  Systems Conference (ITSC)}, 2021, pp. 2816--2821.

\bibitem{kumar2020fisheyedistancenet}
V.~R. Kumar, S.~A. Hiremath, M.~Bach, S.~Milz, C.~Witt, C.~Pinard, S.~Yogamani,
  and P.~M{\"a}der, ``{Fisheyedistancenet: Self-supervised scale-aware distance
  estimation using monocular fisheye camera for autonomous driving},'' in
  \emph{2020 IEEE International Conference on Robotics and Automation (ICRA)},
  2020, pp. 574--581.

\bibitem{kumar2021fisheyedistancenet++}
R.~K. Varun, S.~Yogamani, S.~Milz, and P.~M\"{a}der, ``{FisheyeDistanceNet++:
  Self-Supervised Fisheye Distance Estimation with Self-Attention, Robust Loss
  Function and Camera View Generalization},'' in \emph{Electronic Imaging},
  2021.

\bibitem{kumar2020syndistnet}
V.~Ravi~Kumar, M.~Klingner, S.~Yogamani, S.~Milz, T.~Fingscheidt, and P.~Mader,
  ``{Syndistnet: Self-supervised monocular fisheye camera distance estimation
  synergized with semantic segmentation for autonomous driving},'' in
  \emph{Proceedings of the IEEE/CVF Winter Conference on Applications of
  Computer Vision}, 2021, pp. 61--71.

\bibitem{9626604}
C.~Eising, J.~Horgan, and S.~Yogamani, ``Near-field perception for low-speed
  vehicle automation using surround-view fisheye cameras,'' \emph{IEEE
  Transactions on Intelligent Transportation Systems}, pp. 1--18, 2021.

\bibitem{yahiaoui2019fisheyemodnet}
M.~Yahiaoui, H.~Rashed, L.~Mariotti, G.~Sistu, I.~Clancy, L.~Yahiaoui, and
  S.~Yogamani, ``{FisheyeMODNet}: Moving object detection on surround-view
  cameras for autonomous driving,'' in \emph{Proceedings of the Irish Machine
  Vision and Image Processing (IMVIP)}, 2019, pp. 1--4.

\bibitem{gallagher2021hybrid}
L.~Gallagher, V.~R. Kumar, S.~Yogamani, and J.~B. McDonald, ``A hybrid
  sparse-dense monocular slam system for autonomous driving,'' in \emph{Proc.
  of ECMR}.\hskip 1em plus 0.5em minus 0.4em\relax IEEE, 2021, pp. 1--8.

\bibitem{kumar2018near}
V.~R. Kumar, S.~Milz, C.~Witt, M.~Simon, K.~Amende, J.~Petzold, S.~Yogamani,
  and T.~Pech, ``Near-field depth estimation using monocular fisheye camera: A
  semi-supervised learning approach using sparse lidar data,'' in \emph{CVPR
  Workshop}, vol.~7, 2018, p.~2.

\bibitem{kumar2018monocular}
------, ``{Monocular fisheye camera depth estimation using sparse lidar
  supervision},'' in \emph{2018 21st International Conference on Intelligent
  Transportation Systems (ITSC)}, 2018, pp. 2853--2858.

\bibitem{ruping2022inspect}
S.~R{\"u}ping, E.~Schulz, J.~Sicking, T.~Wirtz, M.~Akila, S.~Gannamaneni,
  M.~Mock, M.~Poretschkin, J.~Rosenzweig, S.~Abrecht \emph{et~al.}, ``Inspect,
  understand, overcome: A survey of practical methods for ai safety,''
  \emph{Deep Neural Networks and Data for Automated Driving: Robustness,
  Uncertainty Quantification, and Insights Towards Safety}, p.~3, 2022.

\bibitem{kim2016fisheye}
H.~Kim, E.~Chae, G.~Jo, and J.~Paik, ``Fisheye lens-based surveillance camera
  for wide field-of-view monitoring,'' in \emph{2015 IEEE International
  Conference on Consumer Electronics (ICCE)}, 2015, pp. 505--506.

\bibitem{schmalstieg2016augmented}
D.~Schmalstieg and T.~Hollerer, \emph{Augmented reality: principles and
  practice}.\hskip 1em plus 0.5em minus 0.4em\relax Addison-Wesley
  Professional, 2016.

\bibitem{maddern20171}
W.~Maddern, G.~Pascoe, C.~Linegar, and P.~Newman, ``1 year, 1000 km: The oxford
  robotcar dataset,'' \emph{The International Journal of Robotics Research},
  vol.~36, no.~1, pp. 3--15, 2017.

\bibitem{liao2022kitti}
Y.~Liao, J.~Xie, and A.~Geiger, ``Kitti-360: A novel dataset and benchmarks for
  urban scene understanding in 2d and 3d,'' \emph{IEEE Transactions on Pattern
  Analysis and Machine Intelligence}, 2022.

\bibitem{sekkat2020omniscape}
A.~R. Sekkat, Y.~Dupuis, P.~Vasseur, and P.~Honeine, ``The omniscape dataset,''
  in \emph{2020 IEEE International Conference on Robotics and Automation
  (ICRA)}.\hskip 1em plus 0.5em minus 0.4em\relax IEEE, 2020, pp. 1603--1608.

\bibitem{yogamani2019woodscape}
S.~Yogamani, C.~Hughes, J.~Horgan, G.~Sistu, P.~Varley, D.~O'Dea,
  M.~Uric{\'a}r, S.~Milz, M.~Simon, K.~Amende \emph{et~al.}, ``Woodscape: A
  multi-task, multi-camera fisheye dataset for autonomous driving,'' in
  \emph{Proceedings of the IEEE/CVF International Conference on Computer
  Vision}, 2019, pp. 9308--9318.

\bibitem{SYNTHIA}
G.~Ros, L.~Sellart, J.~Materzynska, D.~Vazquez, and A.~M. Lopez, ``The synthia
  dataset: A large collection of synthetic images for semantic segmentation of
  urban scenes,'' in \emph{2016 IEEE Conference on Computer Vision and Pattern
  Recognition (CVPR)}, 2016, pp. 3234--3243.

\bibitem{johnson2017driving}
M.~Johnson-Roberson, C.~Barto, R.~Mehta, S.~N. Sridhar, K.~Rosaen, and
  R.~Vasudevan, ``Driving in the matrix: Can virtual worlds replace
  human-generated annotations for real world tasks?'' in \emph{2017 IEEE
  International Conference on Robotics and Automation (ICRA)}.\hskip 1em plus
  0.5em minus 0.4em\relax IEEE, 2017, pp. 746--753.

\bibitem{Richter_2017}
S.~R. Richter, Z.~Hayder, and V.~Koltun, ``Playing for benchmarks,'' in
  \emph{{IEEE} International Conference on Computer Vision, {ICCV} 2017,
  Venice, Italy, October 22-29, 2017}, 2017, pp. 2232--2241.

\bibitem{wang2019apolloscape}
P.~Wang, X.~Huang, X.~Cheng, D.~Zhou, Q.~Geng, and R.~Yang, ``The apolloscape
  open dataset for autonomous driving and its application,'' \emph{IEEE
  transactions on pattern analysis and machine intelligence}, 2019.

\bibitem{Weng2020_AIODrive}
X.~Weng, Y.~Man, J.~Park, Y.~Yuan, D.~Cheng, M.~O'Toole, and K.~Kitani,
  ``{All-In-One Drive: A Large-Scale Comprehensive Perception Dataset with
  High-Density Long-Range Point Clouds},'' \emph{arXiv}, 2021.

\bibitem{carlasimulator}
A.~Dosovitskiy, G.~Ros, F.~Codevilla, A.~Lopez, and V.~Koltun, ``{CARLA}: {An}
  open urban driving simulator,'' in \emph{Proceedings of the 1st Annual
  Conference on Robot Learning}, 2017, pp. 1--16.

\bibitem{kumar2021omnidet}
V.~Ravi~Kumar, S.~Yogamani, H.~Rashed, G.~Sitsu, C.~Witt, I.~Leang, S.~Milz,
  and P.~M{\"a}der, ``Omnidet: Surround view cameras based multi-task visual
  perception network for autonomous driving,'' \emph{IEEE Robotics and
  Automation Letters}, vol.~6, no.~2, pp. 2830--2837, 2021.

\end{thebibliography}
\end{document}